\documentclass[11pt]{article}

\usepackage[final]{acl}

\usepackage{times}
\usepackage{latexsym}
\usepackage{multirow}
\usepackage{amsfonts}
\usepackage{multicol}
\usepackage{multirow}
\usepackage{todonotes}
\usepackage{breqn}
\usepackage{amsmath}
\usepackage{arydshln}
\usepackage{mathtools}
\usepackage{array, makecell}
\usepackage{xcolor}
\usepackage{todonotes}
\usepackage{booktabs}
\usepackage{tcolorbox} 

\usepackage[T1]{fontenc}

\usepackage[utf8]{inputenc}

\usepackage{microtype}

\usepackage{inconsolata}
\usepackage{pifont}
\usepackage{comment}
\usepackage{url}
\usepackage{graphicx}
\usepackage{amsmath}
\usepackage{booktabs}
\usepackage{multirow}
\usepackage{enumitem}
\usepackage{lineno}
\usepackage{xcolor}
\usepackage{subfig}
\usepackage{amsmath}

\usepackage{algorithm}
\usepackage{algorithmic}  
\usepackage{geometry}  
\title{\textit{Temporal Referential Consistency}: \\Do LLMs Favor Sequences Over Absolute Time References?}
\usepackage{amsmath}
\newcommand{\newdata}{\texttt{TEMP-ReCon}}

\newcommand{\newmodel}{\texttt{UnTRaP}}
\newcommand{\newmodelfull}{\texttt{Unifying Temporal Reasoning Pathways}}
\newcommand{\newmodelmulti}{\texttt{$UnTRaP^*$}}

\author{\textbf{Ashutosh Bajpai\textsuperscript{1,2}},  \textbf{Tanmoy Chakraborty\textsuperscript{1}}
\\
\\
\textsuperscript{1}Indian Institute of Technology Delhi, India \\ \textsuperscript{2}MongoDB, Inc., Delhi, India
\\
{
   {\textit {\{eez228482,tanchak\}@ee.iitd.ac.in}}
  }
}

\begin{document}
\maketitle
\begin{abstract}
The increasing acceptance of large language models (LLMs) as an alternative to knowledge sources marks a significant paradigm shift across various domains, including time-sensitive fields such as law, healthcare, and finance. To fulfill this expanded role, LLMs must not only be factually accurate but also demonstrate consistency across temporal dimensions, necessitating robust temporal reasoning capabilities. Despite this critical requirement, efforts to ensure temporal consistency in LLMs remain scarce including noticeable absence of endeavors aimed at evaluating or augmenting LLMs across temporal references in time-sensitive inquiries. In this paper, we seek to address this gap by introducing a novel benchmark entitled temporal referential consistency, accompanied by a resource \newdata\ designed to benchmark a wide range of both open-source and closed-source LLMs with various linguistic contexts characterized by differing resource richness (including English, French, and Romanian). The findings emphasis that LLMs do exhibit insufficient temporal referent consistency. To address this, we propose \newmodel, a reasoning path alignment-based model  that aims to enhance the temporal referential consistency of LLMs. Our empirical experiments substantiate the efficacy of \newmodel\ compared to several baseline models.
\end{abstract}

\section{Introduction}

Temporal reasoning encompasses the understanding of time-related relationships, including time-time, time-event, and event-event correlations \cite{tan-etal-2023-towards}. This capability allows Large Language Models (LLMs) \cite{openai2024gpt4technicalreport,geminiteam2024gemini15unlockingmultimodal} to process and interpret the evolving nature of information, which is essential in fields sensitive to temporal changes, such as finance, healthcare, and legal studies. By enhancing LLMs with robust temporal reasoning abilities, their analytical competencies are significantly improved, facilitating complex tasks such as timelines summarization \cite{10.1145/3626772.3657899}, disease progression \cite{lu-naseem-2024-large}, managing contracts negotiations \cite{narendra-etal-2024-enhancing}, and conducting analysis on historical text \cite{zeng2024histolensllmpoweredframeworkmultilayered}. 

 \begin{figure}[!t]
\centering
\includegraphics[width=0.95\columnwidth]{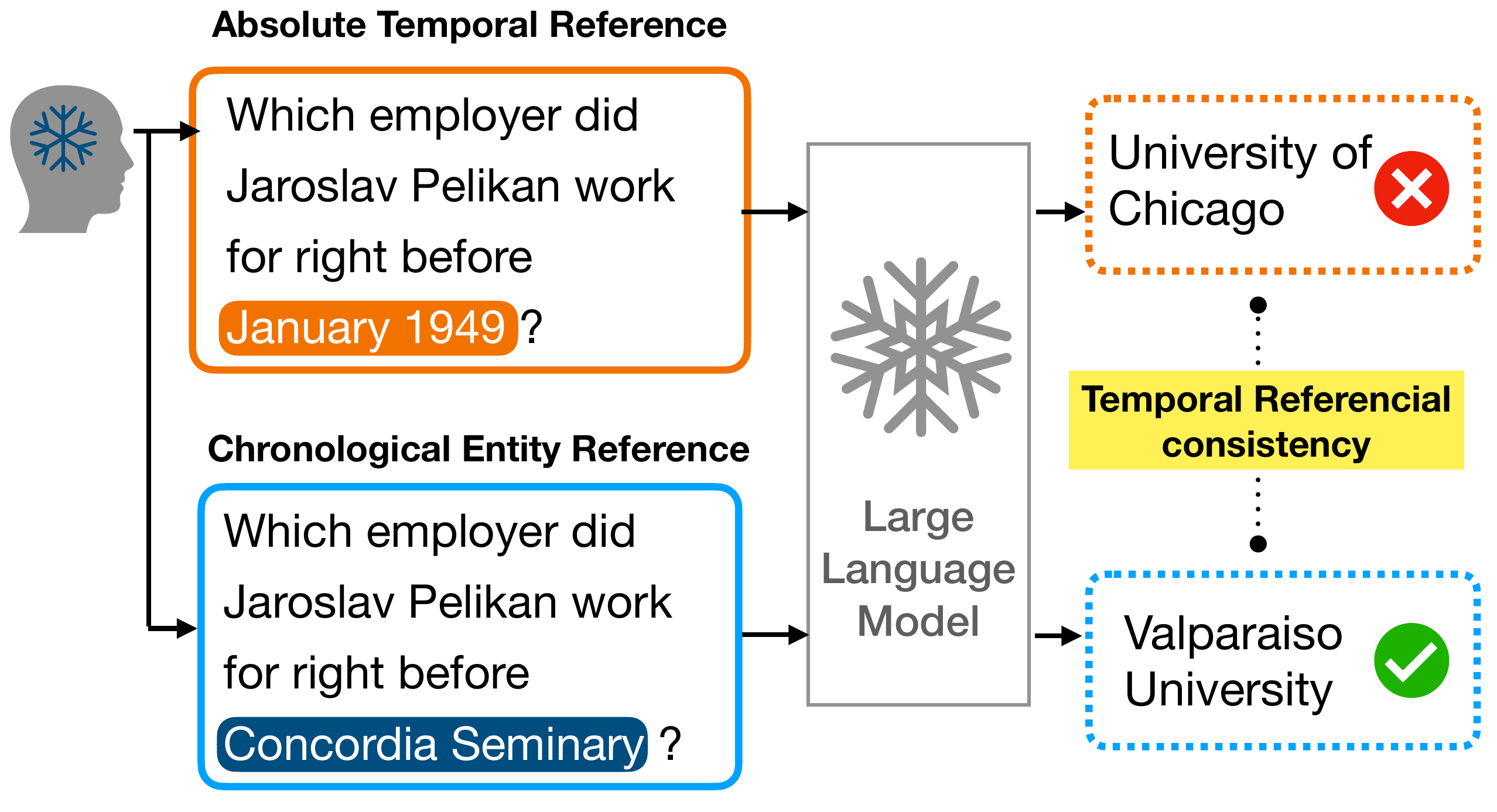}
  \caption{An illustration of temporal referential consistency.}
  \label{fig:maintask}
  \vspace{-5mm}
\end{figure}
\paragraph{Consistency.} Consistency in LLMs ensures reliable and coherent responses across diverse inputs, making them essential for applications such as decision support systems. Knowledge Bases (KBs), built on manually engineered schemas, have historically been used for information retrieval to ensure factual and consistent answers \cite{petroni2019languagemodelsknowledgebases}. Conversely, inconsistent factuality is reported as a prevalent issue in LLMs, especially in autoregressive settings \cite{tam-etal-2023-evaluating}. \citet{ravichander-etal-2020-systematicity} evaluated consistency using paired probes, while \citet{elazar-etal-2021-measuring} expanded this analysis by examining and improving LLM consistency across different types of factual knowledge, thereby defining consistent factuality in LLMs.

\paragraph{Temporal Consistency.}  Recently, \citet{bajpai-etal-2024-temporally} expanded the evaluation of consistent factuality across the temporal axis, marking it as a preliminary attempt in assessing consistent temporal reasoning in LLMs. However, their definition of temporal consistency is limited to surface-form alterations (paraphrasing) of time-sensitive queries, neglecting consistency across various temporal references such as an event or absolute time. To address this gap in the literature, we aim to formalize two research questions in this study.

\begin{itemize}
[itemsep=2pt,nolistsep,topsep=1pt,leftmargin=1em]
    \item \textbf{RQ1:} To what extent do LLMs possess the capability for temporal referential consistency? 
    \item \textbf{RQ2:} How can LLMs be improved to address inadequate temporal referential consistency?
\end{itemize}

As illustrated in Figure \ref{fig:maintask}, temporal referential consistency expects identical responses across absolute (e.g., January 1994) and chronological temporal references, which rely on implicit temporal information (e.g., an event like Concordia Seminary). To this end, we introduce \newdata, a novel multilingual dataset, and expand existing metrics (defined in Section \ref{sec:experiment}) to quantitatively benchmark LLMs for temporal referential consistency across languages. Our study focuses on English, French, and Romanian, chosen for their varied resource richness, with Romanian identified as a low-resource language. Findings indicate that LLMs exhibit inadequate performance regarding temporal referential consistency across languages.

To address this issue, we propose a novel method -- \newmodel\  (\newmodelfull). First, we define two reasoning pathways a model can utilize to arrive at the anticipated answer despite divergent temporal references: event-oriented and time-oriented reasoning pathways (defined in Section \ref{sec:method}). We hypothesize that alignment between reasoning pathways within the LLM's hyperspace will enable the model to consistently respond to variations in temporal references. Therefore, we adopt a supervised instruction-tuning approach for this alignment. While several attempts have been made to identify, evaluate, or optimize reasoning paths \cite{chung2022scalinginstructionfinetunedlanguagemodels,yuan2023scalingrelationshiplearningmathematical}, existing methods lack a unified approach to enhance consistency. Our novel method, \newmodel, addresses this gap by unifying reasoning paths to achieve improved consistency. We compare \newmodel\ with six baselines and demonstrate that it outperforms the best baseline \cite{tan-etal-2023-towards} by 9.06, and 5.47 percentage points, respectively, for temporal referential consistency and temporally referential consistent factuality.

To summarize, our contributions are as follows\footnote{Source code and dataset are available at \url{https://github.com/ab-iitd/untrap}.}:
\begin{itemize}
[itemsep=2pt,nolistsep,topsep=0pt,leftmargin=1em]
    \item We propose \newdata, a multilingual resource to evaluate a new benchmark: temporal referential consistency in LLMs. To achieve this, we build upon the metrics proposed by \citet{bajpai-etal-2024-temporally} for performance benchmarking.
   \item Our findings indicate that LLMs demonstrate superior performance in handling chronological temporal references as opposed to absolute temporal references, thereby resulting in greater temporal referential inconsistency across languages.
   \item We propose \newmodel, which improves the temporal referential consistency of LLMs and empirically outperforms robust baselines across multiple languages.

\end{itemize}
\begin{table}[t]
\centering
\small
\resizebox{.99\columnwidth}{!}{
\begin{tabular}{l@{\hspace{0.3cm}}|@{\hspace{0.2cm}}c|@{\hspace{0.2cm}}c|@{\hspace{0.2cm}}c}
\toprule
& \textbf{Train} & \textbf{Dev} & \textbf{Test} \\
\midrule
Time Range &1015-2023 &634-2023& 1001-2022\\
\#Total Queries &13,014 &4,437 &4,426\\
\midrule
\multicolumn{4}{l}{\textbf{Absolute Temporal Reference Query}}\\
\multicolumn{4}{l}{Which employer did Jaroslav Pelikan work for right }\\
\multicolumn{4}{l}{before January 1949?}\\
\multicolumn{4}{l}{\textbf{Chronological Temporal Reference Query}}\\
\multicolumn{4}{l}{Which employer did Jaroslav Pelikan work for right }\\
\multicolumn{4}{l}{before Concordia Seminary?}\\
\multicolumn{4}{l}{\textbf{Correct Answer}}\\
\multicolumn{4}{l}{Valparaiso University}\\
\multicolumn{4}{l}{\textbf{Time-Oriented Reasoning Pathway}}\\
\multicolumn{4}{l}{Because jaroslav pelikan worked for concordia seminary}\\
\multicolumn{4}{l}{from january 1949 to january 1953, and right before}\\
\multicolumn{4}{l}{january 1949, jaroslav pelikan worked for valparaiso}\\
\multicolumn{4}{l}{university.}\\
\multicolumn{4}{l}{\textbf{Event-Oriented Reasoning Pathway}}\\
\multicolumn{4}{l}{Because jaroslav pelikan worked for concordia seminary}\\
\multicolumn{4}{l}{from january 1949 to january 1953, and right before}\\
\multicolumn{4}{l}{ concordia seminary, jaroslav pelikan worked for}\\
\multicolumn{4}{l}{valparaiso university.}\\
\bottomrule
\end{tabular}
}
\caption{Dataset statistics for \newdata\ along with a representative instance of the dataset.}
\label{tab:datastas}
\vspace{-5mm}
\end{table}
\section{Dataset}\label{sec:dataset}
Despite extensive research in temporal reasoning, no dataset currently exists to specifically support evaluation of LLMs for temporal referential consistency. Among notable datasets, TEMPREASON \cite{tan-etal-2023-towards} emerges as a comprehensive benchmark, offering a multifaceted approach to temporality across an extended temporal period. TEMPREASON is a Q/A style dataset that categorizes temporal reasoning into three distinct types: time-time (L1), time-event (L2), and event-event (L3). It expects time, event, and event as a response from the model when provided with time, time, and event information in a query. Consequently, we sourced this dataset to develop a novel resource for evaluating temporal referential consistency in LLMs, designated as \newdata\footnote{The resource was curated by a male NLP expert, within the 30-40 age bracket.} (Temporal Referential Consistency). We select the instances of event-event category (L3) relations from TEMPREASON due to its provision of relations among events alongside the temporal periods associated with them. An event-event temporal query involves identifying an answer-object chronologically related to a query-object, based on a subject-relation-object triplet and directional cues like "before" or "after."
\begin{figure*}[!t]
\centering
\includegraphics[width=0.99\textwidth]{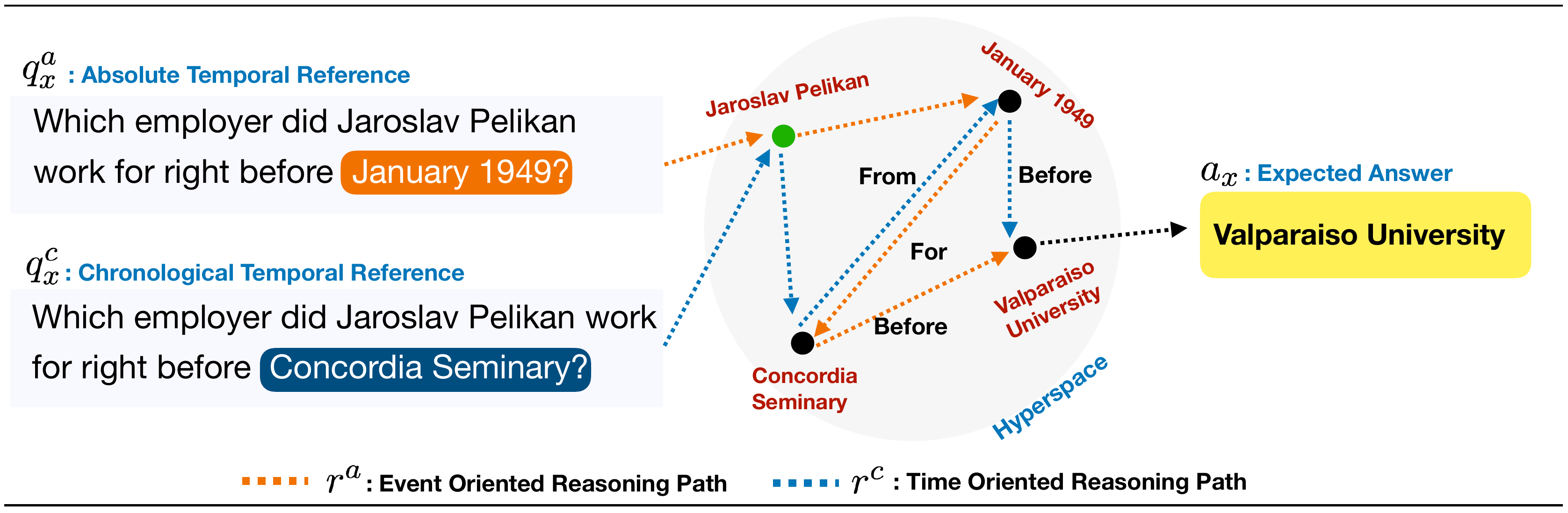}
  \caption{The architectural formulation of \newmodelfull\ (\newmodel). $r^a$: an event-oriented reasoning path that sequences time-event:event-event to address absolute temporal reference queries. $r^c$: a time-oriented reasoning path that follows event-time:time-event for chronological temporal reference queries.}
  \label{fig:modelfig}
  \vspace{-3mm}
\end{figure*}
\paragraph{\newdata.} An instance of \newdata\ comprises five fields: absolute temporal reference query, chronological temporal reference query, correct answer, time-oriented reasoning pathway, and event-oriented reasoning pathway. To construct the chronological temporal reference-based query, we utilize the default L3 question from the dataset, subsequently replacing the query event with the appropriate absolute temporal reference corresponding to the time associated with the event to obtain chronological temporal reference query. Additionally, we devise the event and time-oriented reasoning paths based on information provided under the field \textit{fact\_context}, present in source data. \textit{fact\_context} states all the temporal facts for a query. Algorithm \ref{algo:data_construction} presents details on data construction.

\noindent\textbf{Multilingual Expansion.} Further, we extend the dataset to non-english languages to support the evaluation of multilingual temporal referential consistency. Inspired by \citet{Bajpai_Chakraborty_2025}, we select French and Romanian due to their varying levels of resource availability in comparison to English. Romanian and French possess 98.3\% and 78\% fewer speakers than English, respectively. We utilize the T5 model \cite{raffel2023exploringlimitstransferlearning} to facilitate the automatic translation from English to Romanian and French. We rely on the source dataset for the data quality standard. Additionally, we report the mean Translation Success Rate (TSR) of 99.4\% and the mean Back-Translation Accuracy (BTA) of 71.49\% and 45.49\%, based on BERT-Score and ChrF++ \cite{popovic-2015-chrf}, respectively, to evaluate translation quality. To report inter-translator agreement, we employed NLLB-200 \cite{nllbteam2022languageleftbehindscaling} alongside T5, achieving mean translation agreement scores of 93.75 and 62.48 as measured using BERTScore and BLEU-3 based metrics. Dataset statistics, along with a representative instance, is detailed in Table \ref{tab:datastas} (Reasers can refer to Appendix \ref{app:dataquality} for more details on proposed data).
  
\begin{algorithm}[!t]
\small
\caption{Data Construction Approach}  
\begin{algorithmic}[1]
\STATE \textbf{Step 1: Selection of Data Points}  
\STATE Select data points from the event-event category of TEMPREASON, leveraging its provision of event relations and temporal periods.  
  
\STATE \textbf{Step 2: Adding 'right' Keyword}  
\STATE Add the keyword "right" before the reference terms "before/after" in the query to restrict probing to a one-to-one setting and access the immediate entity before or after the provided temporal reference.  
  
\STATE \textbf{Step 3: Generating Chronological Temporal Reference Query}  
\STATE Transform the query using the added keyword "right" to produce chronological temporal reference queries.  
  
\STATE \textbf{Step 4: Generating Absolute Temporal Reference Query}  
\begin{itemize}
[itemsep=2pt,nolistsep,topsep=0pt,leftmargin=1em]
    \item Extract the reference entity from the query using reference terms 'before/after'.
    \item Search the `fact\_context` field in TEMPREASON for the extracted entity to retrieve its "from" and "to" time periods.  
    \item Based on the keyword ("before/after"), select the "from" or "to" time for the entity.  
    \item Replace the reference entity in the query with the absolute time to create the absolute temporal reference query.  
\end{itemize}  
  
\STATE \textbf{Step 5: Constructing Reasoning Pathways}  
\STATE Build two reasoning pathways:  
  
\STATE \textbf{Step 5.1: Time-Oriented Reasoning Pathway}  
\begin{itemize}
[itemsep=2pt,nolistsep,topsep=0pt,leftmargin=1em]
    \item Extract the absolute time reference from the absolute temporal reference query.  
    \item Search the `fact\_context` for facts corresponding to $t+1$ or $t-1$ based on "after" or "before" type of query.  
    \item Extract the fact containing the correct answer, remove "from" and "to" time information from it, and concatenate using the phrase "right before/after [absolute time reference]."  
    \item Prefix with the keyword "because" to construct the final time-oriented reasoning pathway.  
\end{itemize}  
  
\STATE \textbf{Step 5.2: Event-Oriented Reasoning Pathway}  
\begin{itemize}
[itemsep=2pt,nolistsep,topsep=0pt,leftmargin=1em]
    \item Extract the event reference from the chronological temporal reference query.  
    \item Search the `fact\_context` for facts corresponding to $t+1$ or $t-1$ based on "after" or "before" type of query. 
    \item Extract the fact containing the correct answer, remove "from" and "to" time information, and concatenate it using the phrase "right before/after [chronological temporal reference]."  
    \item Prefix with the keyword "because" to construct the final event-oriented reasoning pathway.  
\end{itemize}  
  
\end{algorithmic} 
\label{algo:data_construction}
\end{algorithm} 

\section{Proposed Methodology}\label{sec:method}

\subsection{Preliminaries}
Let us consider a dataset $D$ that comprises $M$ pairs of referential paraphrases of temporal queries, denoted as ($q^a_x, q^c_x$), along with an associated answer $a_x$. In this context, $q^a_x$ represents the temporal query with absolute temporal reference, while $q^c_x$ represents the temporal query with chronological temporal reference, where $x \in M$. Within a classical supervised fine-tuning framework, a model $\Theta$ is subjected to the loss objective $L_{sft}$, as articulated in Equation \ref{eq:eq_sft}. The primary objective of this framework is to jointly maximize the probability of obtaining $a_x$ given either $q^a_x$ or $q^c_x$.
\vspace{-3mm}
\begin{equation}
\begin{split}
\mathcal{L}_{sft} = - \sum_{x=1}^{M} log P(a_x | \, q^a_x) +  log P(a_x | \, q^c_x)   
   \end{split}
   \label{eq:eq_sft}
\end{equation}
\citet{bajpai-etal-2024-temporally}  extended the earlier formulation to a multi-task (MT) fine-tuning framework. This approach demonstrates that the inclusion of an explicit consistency task significantly influences task transfer, thereby enhancing the temporally consistent factuality within the model. The model, denoted as $\Theta,$ is subject to joint fine-tuning through an MT objective that encompasses both factuality and consistency. The loss objective pertaining to the consistency task, denoted as $L_{con}$, is presented in Equation \ref{eq:eq_con}. Consistency is framed as a binary generative task, wherein the objective is to maximize the probability of the ground truth class labels $True$ and $False$, corresponding to two positive and antagonistic time-sensitive referential paraphrases, respectively. Here, the class label $c$ is defined within the set $\{True, False\}$.
\vspace{-2mm}
\begin{equation}
\begin{split}
\mathcal{L}_{con} = - \sum_{x=1}^{N} log P(c_x \,| \, (q^a_x, q^c_x))   
   \end{split}
   \label{eq:eq_con}
\end{equation}
Therefore, the final loss objective of the MT approach, $L_{total}$,  is delineated in Equation \ref{eq:eq_mt}. In this formulation, $L_{fact}$ represents the loss objective for the primary task of temporal factuality, consistent with Equation \ref{eq:eq_sft}. Furthermore, they have introduced consistent time-sensitive reinforcement learning (CTSRL) to enhance the model $\Theta$ by implementing an independent reward mechanism for both tasks within the MT approach.
\begin{equation}
\begin{split}
\mathcal{L}_{total} =  \mathcal{L}_{fact} + \, \mathcal{L}_{con}
\end{split}
\label{eq:eq_mt}
\end{equation}

\subsection{\newmodelfull\ (\newmodel)} Thus far, the objective presents simplistic assumptions that establish a direct association between the answer $a_x$ and the query $q_x$. However, it neglects to assess or enhance the logical deductions-- termed time-sensitive reasoning pathways-- that link an answer to a query. Such deductions may also be multi-step in nature. To this end, as illustrated in Figure \ref{fig:modelfig}, we define two time-sensitive reasoning pathways, $r^a$ and $r^c$, for accessing the answer $a_x$ given queries $q^a_x$ and $q^c_x$. The first time-sensitive reasoning pathway, $r^a$, given a query with absolute temporal reference, adheres to the event-oriented order of \textit{time-event:event-event}. For instance, consider the query: \textit{Which employer did Jaroslav Pelikan work for right before January 1949?} This query provides an absolute temporal reference (January 1949) to ascertain the correct answer, an event (Valparaiso University). Consequently, a reasoning pathway may first lead to another event, denoted as event (Concordia Seminary), utilizing the specified time (January 1949) and then proceed from event (Concordia Seminary) to the anticipated event (Valparaiso University). Thus, $r^a$ for the given example can be articulated as: \textit{Jaroslav Pelikan worked for Concordia Seminary from January 1949 to January 1953, and prior to Concordia Seminary, Jaroslav Pelikan was employed by Valparaiso University.}
\begin{equation}
   P^a = \arg\max_{a_x \in A} p\,(a_x, \,r^a_x \,| \ q^a_x)
   \label{eq:maxlikelihood1}
\end{equation}
\begin{equation}
   P^c = \arg\max_{a_x \in A} p\,(a_x, \,r^c_x \,| \ q^c_x)
   \label{eq:maxlikelihood2}
\end{equation}
Whereas another time-sensitive reasoning pathway $r^c$, given a query with chronological temporal reference, adheres to the time-oriented sequence of \textit{event-time:time-event} order. Consider the same case within a chronological temporal reference setting. The query is - \textit{Which employer did Jaroslav Pelikan work for immediately prior to Concordia Seminary?}, where the chronological temporal reference, an event (Concordia Seminary), is utilized to ascertain the correct answer, which is another event (Valparaiso University). In this scenario, we first identify an absolute temporal reference, time (January 1949), corresponding to the given event, and subsequently follows the pathway from this time to the expected event (Valparaiso University). Thus, the reasoning pathway $r^c$ for the given example can be articulated as \textit{Jaroslav Pelikan worked for Concordia Seminary from January 1949 to January 1953, and immediately prior to January 1949, he was employed by Valparaiso University.}
\vspace{-2mm}
\begin{equation}
\begin{split}
\mathcal{L}_{\newmodel} = - \sum_{x=1}^{M} log P^a (&a_x, \,r^a_x \,| \ q^a_x) \, + \\&log P^c(a_x, \,r^c_x \,| \ q^c_x)
   \end{split}
   \label{eq:eq_untrap}
\end{equation}

\begin{table*}[t]
\centering
\small
\resizebox{.99\textwidth}{!}{
\begin{tabular}{l|l|c|c|c|c|c|c|c|c}
\toprule
\textbf{Model}	&\textbf{Method}	&\multicolumn{6}{c|}{\textbf{Temp-Ref-Fact}}&	\textbf{Temp-Ref-Cons}&	\textbf{Temp-Ref-Cons-Fact}\\
\cline{3-8}
& &\multicolumn{3}{c|}{\textbf{EM}} &\multicolumn{3}{c|}{\textbf{F1.}} & & \\
\cline{3-8}
& &\textbf{CTR}&\textbf{ATR}&\textbf{Dev.} &\textbf{CTR}&\textbf{ATR}&\textbf{Dev.} & & \\
\midrule
\multicolumn{10}{c}{\textit{Open-Source LLMs}}\\
\midrule

\multirow{3}*{LLaMA3.1-8B}	&ICL&7.95	&5.01	&-2.94	&18.82	&13.40	&-5.42	&26.77	&2.76\\
&Semantic ICL&12.94	&10.23	&-2.71	&20.96	&19.76	&-1.20	&20.70	&6.82\\
&Semantic CoT&17.84	&13.51	&-4.34	&24.58	&23.10	&-1.48	&23.32	&10.42\\
	\midrule						
\multirow{3}*{Mistral-v1}	&ICL&4.27	&3.29	&-0.98	&16.61	&10.15	&-6.46	&18.53	&1.81\\
&Semantic ICL&11.95	&7.79	&-4.16	&22.50	&18.25	&-4.25	&18.12	&5.08\\
&Semantic CoT&17.21	&12.29	&-4.92	&24.84	&21.76	&-3.08	&22.10	&9.67\\
			\midrule				
\multirow{3}*{Vicuna-7b-v1.5}&ICL&2.59	&0.88	&-1.71	&12.70	&6.79	&-5.91	&14.21	&0.41\\
&Semantic ICL&7.36	&5.28	&-2.08	&15.95	&13.61	&-2.34	&11.32	&3.37\\
&Semantic CoT&11.72	&5.17	&-6.55	&17.47	&12.52	&-4.95	&14.71	&4.20\\
	\midrule						
\multirow{3}*{Bloomz-7b1}&ICL&1.35	&0.65	&-0.70	&8.08	&5.84	&-2.24	&38.84	&0.23\\
&Semantic ICL&3.14	&2.80	&-0.34	&7.49	&8.51	&1.02	&23.81	&0.93\\
&Semantic CoT&7.72	&4.08	&-3.64	&14.85	&12.09	&-2.76	&12.36	&2.42\\
\midrule
\multicolumn{10}{c}{\textit{Closed-Source LLMs}}\\
\midrule
\multirow{3}*{Gemini 2.5 Pro}&ICL&36.4&25.4&-11.0&47.5&39.0&-8.5&33.8 &19.2 \\
&Semantic ICL&38.4&29.4&-9.0&49.2&38.2&-11.0 &35.2 &23.4\\
&Semantic CoT&40.2&36.4&-3.8&49.37&48.18&-1.19&42.2&29.0\\
\bottomrule
\end{tabular}
}
\caption{Default performance of various LLMs on temporal referential consistency across methods -- ICL, Semantic ICL, and Semantic CoT, for English test set. We choose a 3-shot setting across methods. Temp-Ref-Fact (Dev.): temporal factuality deviation across references (in \%), Temp-Ref-Cons: temporal referential consistency (in \%), Temp-Ref-Cons-Fact: temporally referential consistent factuality (in \%). EM is an exact match-based metric. CTR and ATR represents chronological and absolute temporal references.}
\label{tab:res_across_models}
\end{table*}
The objective is to establish unified time-sensitive reasoning pathways across temporal references. It is anticipated that the absolute temporal reference-based query adheres to the event-oriented path, while the chronological temporal reference-based query aligns with the time-oriented path. This cross alignment between the two pathways aims to enhance referential consistency and maximize temporal factuality. The likelihood estimation for absolute temporal reference and chronological temporal reference is denoted as $P^a$ and $P^c$, as presented in Equations \ref{eq:maxlikelihood1} and \ref{eq:maxlikelihood2}, respectively. Meanwhile, $A$ represents the entire vocabulary space encompassing the anticipated answer $a_x$. Consequently, this leads to the final loss objective $L_{\newmodel}$ detailed in Equation \ref{eq:eq_untrap}.

\section{Experiments}
\subsection{Experimental Setting}\label{sec:experiment}

We primarily utilize LLaMA3.1-8B \cite{grattafiori2024llama3herdmodels} for all experimental procedures. We employ LoRA-based supervised instruction-tuning (IT) \cite{10.5555/3600270.3602281,hu2021loralowrankadaptationlarge} for implementing \newmodel\ on the base language, English. Furthermore, we finetune \newmodelmulti\ using training instances across multiple languages. We utilize a test set to benchmark all the results presented in this study. Details including prompts presented in appendices \ref{app:extended} and \ref{App:prompts}.

\begin{table*}[!t]
\centering
\small
\resizebox{.99\textwidth}{!}{
\begin{tabular}{l|l|c|c|c|c|c|c|c|c}
\toprule
\textbf{Model}	&\textbf{Method}	&\multicolumn{6}{c|}{\textbf{Temp-Ref-Fact}}&	\textbf{Temp-Ref-Cons}&	\textbf{Temp-Ref-Cons-Fact}\\
\cline{3-8}
& &\multicolumn{3}{c|}{\textbf{EM}} &\multicolumn{3}{c|}{\textbf{F1.}} & & \\
\cline{3-8}
& &\textbf{CTR}&\textbf{ATR}&\textbf{Dev.} &\textbf{CTR}&\textbf{ATR}&\textbf{Dev.} & & \\
\midrule
\multirow{3}*{English}	&ICL&7.95	&5.01	&-2.94	&18.82	&13.40	&-5.42	&26.77	&2.76\\
&Semantic ICL&12.94	&10.23	&-2.71	&20.96	&19.76	&-1.20	&20.70	&6.82\\
&Semantic CoT&17.84	&13.51	&-4.34	&24.58	&23.10	&-1.48	&23.32	&10.42\\
\midrule			
\multirow{3}*{French}	&ICL&2.14&	3.02&	0.88&	10.95&	9.47	&-1.48&	10.94&	0.95\\
&Semantic ICL	&8.58&	8.33	&-0.25	&17.77	&16.53	&-1.24&	10.21&	4.13\\
&Semantic CoT	&7.68&	6.91&	-0.77	&19.95	&17.90	&-2.05&	19.30&	3.23\\
\midrule			
\multirow{3}*{Romanian}&ICL&2.32	&2.28&	-0.04&	7.07&	6.79&	-0.28&	8.65&	0.77\\
&Semantic ICL	&11.56	&8.35&	-3.21&	16.85&	14.54&	-2.31&	12.18&	6.57\\
&Semantic CoT	&12.78	&9.55&	-3.23&	18.68&	16.42&	-2.26&	20.40&	7.37\\
\bottomrule
\end{tabular}
}
\caption{Default performance of LLaMA3.1-8B on temporal referential consistency across languages- English, French, and Romanian. We choose a 3-shot setting across various methods.}
\label{tab:res_across_lang}
\vspace{-5mm}
\end{table*}

\paragraph{Baselines.} The experimental baselines include several established approaches. In-Context Learning (ICL) \cite{NEURIPS2020_1457c0d6}, Semantic ICL \cite{liu-etal-2022-makes}, Semantic Chain-of-Thoughts Reasoning (Semantic CoT) builds on standard CoT \cite{NEURIPS2022_9d560961}, SFT Model \cite{10.5555/3600270.3602281}, SFT + TSRL \cite{tan-etal-2023-towards}, and CoTSeLF \cite{bajpai-etal-2024-temporally} (see Appendix \ref{app:extended} for details.)

\paragraph{Evaluation Metrics.} We employ and refine the definition of evaluation metrics established by \citet{elazar-etal-2021-measuring,bajpai-etal-2024-temporally} to evaluate the model's temporal referential consistency. The first metric, referred to as \textit{Temporal Referential Factual Deviation}, quantifies the disparity in temporal factuality between queries based on absolute and chronological temporal references as depicted in Equation \ref{eq:TRFD}. This evaluation is grounded in two established metrics: Exact Match (EM), presented in Equation \ref{eq:EM}, and similarly for F1-Score (F1) to measure partial correctness, which assess temporal factuality across temporal references. Here, (*) is a placeholder for absolute or chronological temporal reference and $a'$ is the generated answer. The subsequent metric, termed \textit{Temporal Referential Consistency}, employs a binary estimation -- assigning a value of one or zero -- to determine whether the model's responses to a given pair of queries remain identical or differ across temporal references, presented in Equation \ref{eq:trc}.

\begin{equation}
EM^*_x = 
\begin{cases}
1, & \text{if } {a_x} = {{a'}^*_x} \\
0, & \text{if } {a_x} \neq {{a'}^*_x}
\end{cases}
\label{eq:EM}
\end{equation}
Temporal Referential Factual Deviation
\begin{equation}
 = \frac{1}{M}(\sum_{x=1}^{M} em_x^{a} - \sum_{x=1}^{M}em_x^{c})
\label{eq:TRFD}
\end{equation}
Temporal Referential Consistency
\begin{equation}
=  \frac{1}{M} \sum_{x=1}^{M}
\begin{cases}  
1, & \text{if } {a'}_x^{a} = {a'}_x^{c} \\ 
0, & \text{if } {a'}_x^{a} \neq {a'}_x^{c} 
\end{cases}
\label{eq:trc}
\end{equation}  
The third and final metric, designated as \textit{Temporally Referential Consistent Factuality}, serves as a composite measure of the preceding two metrics, establishing a more stringent criterion for model responses to be both temporally consistent and factually accurate across absolute and chronological temporal references, presented in Equation \ref{eq:trcf}.

Temporal Referential Consistent
Factuality
\vspace{-2mm}
\begin{equation} 
=  \frac{1}{M} \sum_{x=1}^{M}
\begin{cases}  
1, & \text{if } {a'}_x^{a} = {a'}_x^{c} \\    
   & \text{and if } {a'}_x^{a} = a_x^{c} \\    
0, & \text{if } {a'}_x^{a} \neq {a'}_x^{c}
\end{cases}  
\label{eq:trcf}
\end{equation}  

\subsection{Experimental Results}
\begin{table*}[!ht]
\centering
\small
\resizebox{.99\textwidth}{!}{
 \begin{tabular}{l|c|c|c|c|c|c|c|c}
\toprule
\textbf{Method}	&\multicolumn{6}{c|}{\textbf{Temp-Ref-Fact}}&	\textbf{Temp-Ref-Cons}&	\textbf{Temp-Ref-Cons-Fact}\\
\cline{2-7}
 &\multicolumn{3}{c|}{\textbf{EM}} &\multicolumn{3}{c|}{\textbf{F1.}} & & \\
\cline{2-7}
&\textbf{CTR}&\textbf{ATR}&\textbf{Dev.} &\textbf{CTR}&\textbf{ATR}&\textbf{Dev.} & & \\
\midrule
ICL&7.95	&5.01	&-2.94	&18.82	&13.40	&-5.42	&26.77	&2.76\\
Semantic ICL&12.94	&10.23	&-2.71	&20.96	&19.76	&-1.20	&20.70	&6.82\\
Semantic ICL&17.84	&13.51	&-4.34	&24.58	&23.10	&-1.48	&23.32	&10.42\\

SFT (IT)	&13.30&	11.99&	-1.31	&21.26&	20.89&	-0.37&	32.99&	8.40\\
IT + TSRL $\uparrow$ &14.03&	11.54&	-2.49	&22.22&	21.82&	-0.39&	33.57&	8.20\\
MT-IT&8.40&	7.77&	-0.63	&22.04&	21.46&	-0.57&	24.22&	3.84\\
CoTSeLF (MT-IT + CTSRL) $\Uparrow$&9.69&	8.54&	-1.15	&21.68&	21.14&	-0.53&	25.94&	4.7\\
\textbf{\newmodel} &17.30&	16.85&	\textbf{-0.44}	&25.01&	25.00&	\textbf{-0.01}&	\textbf{42.63}&	\textbf{13.67}\\

\midrule
$\Delta_{\newmodel\ - \uparrow}$& \textcolor{blue}{$3.27\uparrow$}& \textcolor{blue}{$5.31\uparrow$}& \textcolor{blue}{$2.04\uparrow$} & \textcolor{blue}{$2.79\uparrow$}& \textcolor{blue}{$3.18\uparrow$}&\textcolor{blue}{$0.39\uparrow$} & \textcolor{blue}{$9.06\uparrow$} & \textcolor{blue}{$5.47\uparrow$} \\
$\Delta_{\newmodel\ - \Uparrow}$& \textcolor{blue}{$7.61\uparrow$}& \textcolor{blue}{$8.31\uparrow$}& \textcolor{blue}{$0.70\uparrow$}& \textcolor{blue}{$3.33\uparrow$}& \textcolor{blue}{$3.86\uparrow$}& \textcolor{blue}{$0.53\uparrow$} & \textcolor{blue}{$16.69\uparrow$} & \textcolor{blue}{$8.97\uparrow$} \\
\bottomrule
\end{tabular}
}
\caption{Experimental results for \newmodel\ across metrics -- Temp-Ref-Fact-Dev: temporal factuality deviation across references (in \%), Temp-Ref-Cons: temporal referential consistency (in \%), Temp-Ref-Cons-Fact: temporally referential consistent factuality (in \%), in comparison to multiple baselines on test data with LLaMA3.1-8B. We opt for a 3-shot setting across contextual methods. $\Delta_{\newmodel\ - \uparrow}$: improvement by \newmodel\ over TSRL, and $\Delta_{\newmodel\ - \uparrow}$: improvement by \newmodel\ over CoTSeLF.}
\label{tab:res_main}
\vspace{-3mm}
\end{table*}
\begin{table}[!t]
\centering
\small
\resizebox{.99\columnwidth}{!}{
\begin{tabular}{l|l|c|c|c|c}
\toprule
\textbf{Language}&\textbf{Methods}&\textbf{M1}&\textbf{M2}&\textbf{M3}&\textbf{M4}\\
\midrule
French&	Semantic CoT&	-0.77&	-2.05&	19.30&	3.23\\
&	\newmodelmulti	&0.27	&0.55&	31.99	&10.96\\
\midrule
Romanian&	Semantic CoT&	-3.23&	-2.26	&20.40&	7.37\\
&	\newmodelmulti	&-0.39	&0.5	&36.42&	11.5\\
\midrule
English&	Semantic CoT&	-4.34&	-1.48&	23.32&	10.42\\
	&\newmodelmulti&	0.54	&0.74	&39.36&	12.83\\
\midrule
$\overline{\Delta}$& & \textcolor{blue}{$2.38\uparrow$}& \textcolor{blue}{$1.33\uparrow$} & \textcolor{blue}{$14.91\uparrow$} & \textcolor{blue}{$4.75\uparrow$} \\
\bottomrule
\end{tabular}
}
\caption{\newmodelmulti\ over Semantic CoT baseline across languages and metrics -- M1:Temp-Ref-Fact-Dev (EM), M2:Temp-Ref-Fact-Dev (F1), M3:Temp-Ref-Cons, and M4: Temp-Ref-Cons-Fact with LLaMA3.1-8B.}
\label{tab:res_mul}
\vspace{-3mm}
\end{table}
\paragraph{Temporal Referential Consistency Across LLMs.} We start with benchmarking temporal referential consistency across distinct contemporary LLMs, including the multilingual model -- LLaMA3.1-8B, the English-dominant instruction-tuned model Vicuna \cite{zheng2023judgingllmasajudgemtbenchchatbot}, the French and German fluent Mistral \cite{jiang2023mistral7b}, and the cross-lingual specialized Bloomz \cite{muennighoff-etal-2023-crosslingual}. We utilize three prompting strategies: ICL, Semantic ICL, and Semantic CoT. Table \ref{tab:res_across_models} shows that all models exhibit poor performance, with metrics ranging from -0.34\% to -4.34\% for EM-based temporal referential factual deviation, 11.32\% to 38.84\% for temporal referential consistency, and 0.23\% to 10.42\% for temporally referential consistent factuality. Notably, models perform better on chronological reference queries than on absolute temporal references. Additionally, prompting strategies significantly influence performance, with Semantic CoT enhancing temporally referential consistent factuality while exacerbating the performance gap for EM-based temporal factuality. We use an English test set for this experiment. Additionally, we employ, Gemini-2.5-Pro, one of the most contemporary benchmarks in closed-source LLMs. We observe that it also demonstrates suboptimal performance regarding temporal references despite significant improvements over counterparts in open-source category.

\paragraph{Suboptimal Performance Across Languages.}
Next, we benchmark temporal referential consistency across distinct languages
. We employ LLaMA3.1-8B and consider English, French, and Romanian for this experiment across three prompting strategies: ICL, Semantic ICL, and Semantic CoT. Table \ref{tab:res_across_lang} indicates that the model demonstrates suboptimal performance, with metrics ranging from -0.04\% to -4.34\% for EM-based temporal referential factual deviation, 8.65\% to 26.77\% for temporal referential consistency, and 0.77\% to 10.42\% for temporally referential consistent factuality across languages. The models perform better on chronological reference queries than on absolute temporal references across languages. Additionally, there appears to be no significant correlation between the model's temporal referential consistency traits and the resource richness of a language, potentially attributable to the model's extremely poor performance across languages.

\subsubsection{Improvements with \newmodel}
The exhaustive comparison of \newmodel\ with baseline methods demonstrates the efficacy of aligning event- and time-oriented reasoning pathways alongside corresponding temporal references, particularly in contrast to the multi-task reinforcement learning-based method, CoTSeLF, as illustrated in Table \ref{tab:res_main}. Notably, \newmodel\ achieves enhancements of 16.69 and 8.97 percentage points over CoTSeLF for temporal referential consistency and temporal referential consistent factuality, respectively. Furthermore, \newmodel\ exhibits improvements of 9.06 and 5.47 percentage points over the temporal factuality-based baseline, TSRL for temporal referential consistency and temporal reference consistent factuality, respectively. Consistent improvements in temporal factuality across temporal references demonstrate that \newmodel\ enhances consistency without compromising factual accuracy. English variant of \newdata\ is used for this experiment.
\begin{table}[!t]
\centering
\small
\resizebox{.99\columnwidth}{!}{
\begin{tabular}{l|l|c|c|c|c}
\toprule
\textbf{Model}&\textbf{Methods}&\textbf{M1}&\textbf{M2}&\textbf{M3}&\textbf{M4}\\
\midrule

LLaMA3.1-8B&	Semantic CoT&	-4.34& -1.48 &23.32 &10.42\\
&	\newmodel	&\textbf{-0.45}& \textbf{-0.01}& \textbf{42.63} &\textbf{13.67}\\
\midrule
LLaMA2-13B-Chat&	Semantic CoT&-2.03	&-0.91	&13.1&	2.55\\
&	\newmodel	&\textbf{-1.00}	&\textbf{0.26}&	\textbf{43.81}&	\textbf{12.92}\\

\bottomrule
\end{tabular}
}
\caption{\newmodel\ over Semantic CoT baseline across base-models and  metrics -- M1:Temp-Ref-Fact-Dev (EM), M2:Temp-Ref-Fact-Dev (F1), M3:Temp-Ref-Cons, and M4: Temp-Ref-Cons-Fact for English.}
\label{tab:res_gen}
\vspace{-2mm}
\end{table}
\paragraph{Generalization Across Languages.} Results presented in Table \ref{tab:res_mul}, demonstrate that \newmodelmulti\ consistently outperforms the Semantic CoT baseline across all metrics and languages -- French, Romanian, and English. Notably, \newmodelmulti\ exhibits substantial improvements in low-resource languages such as Romanian, achieving temporal referential consistency score of 36.42 compared to 20.40 from the Semantic CoT baseline -- a significant gain of 14.91. Similarly, across high-resource languages like English, \newmodelmulti\ showcases robust generalization, achieving temporal referential consistent accuracy improvement of 4.75.
\begin{figure}[!t]
\centering
\includegraphics[width=0.95\columnwidth]{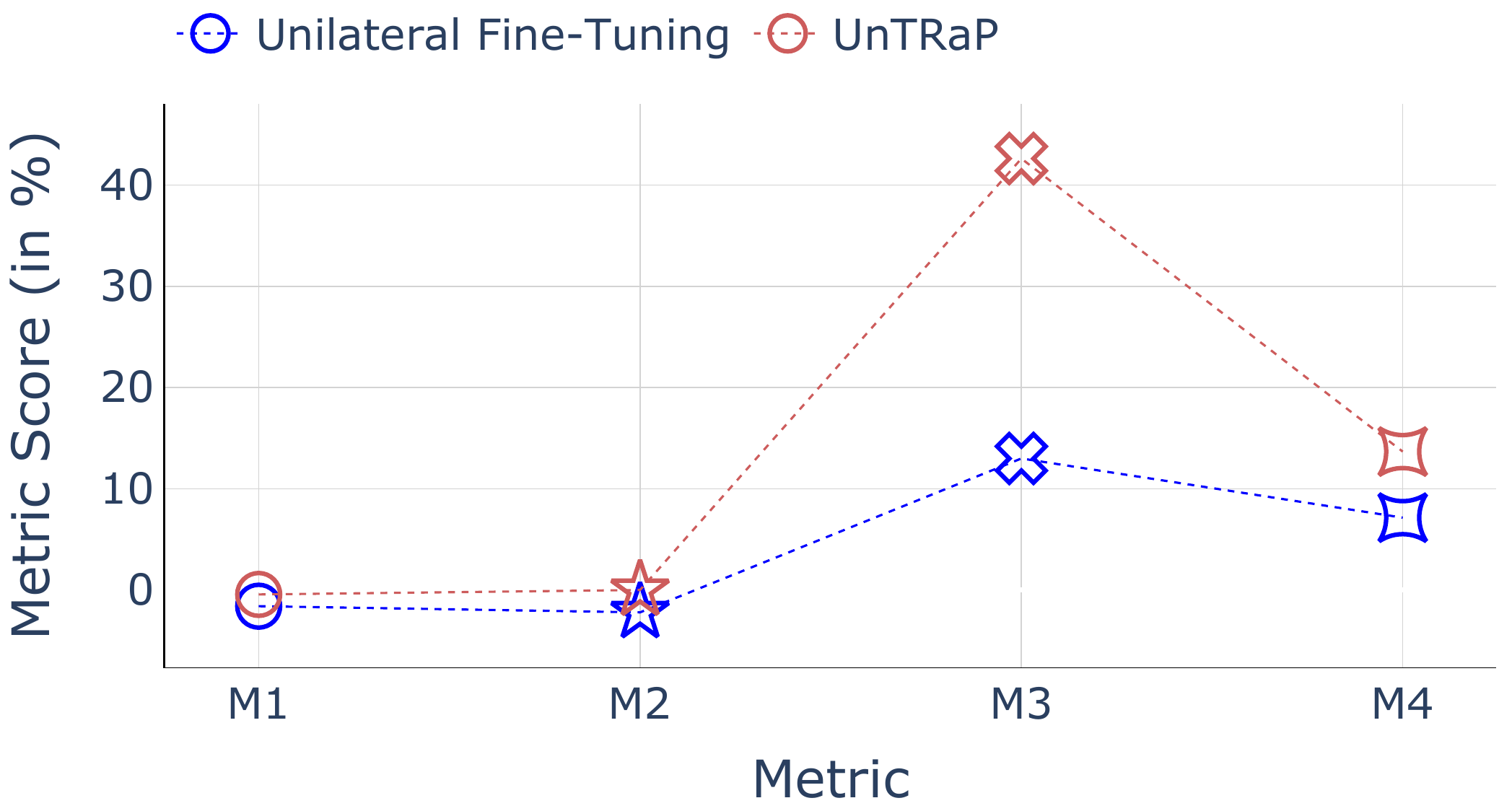}
  \caption{A comparison between \newmodel\ with unilateral (event-oriented) reasoning path fine-tuning utilizing only absolute temporal reference training instances. M1: Temp-Ref-Fact-Dev[EM], M2:Temp-Ref-Fact-Dev[F1], M3: Temp-Ref-Cons, M4: Temp-Ref-Cons-Fact.
  }
  \label{fig:unilateral}
\end{figure}
 \begin{figure}[!t]
\centering
\includegraphics[width=0.95\columnwidth]{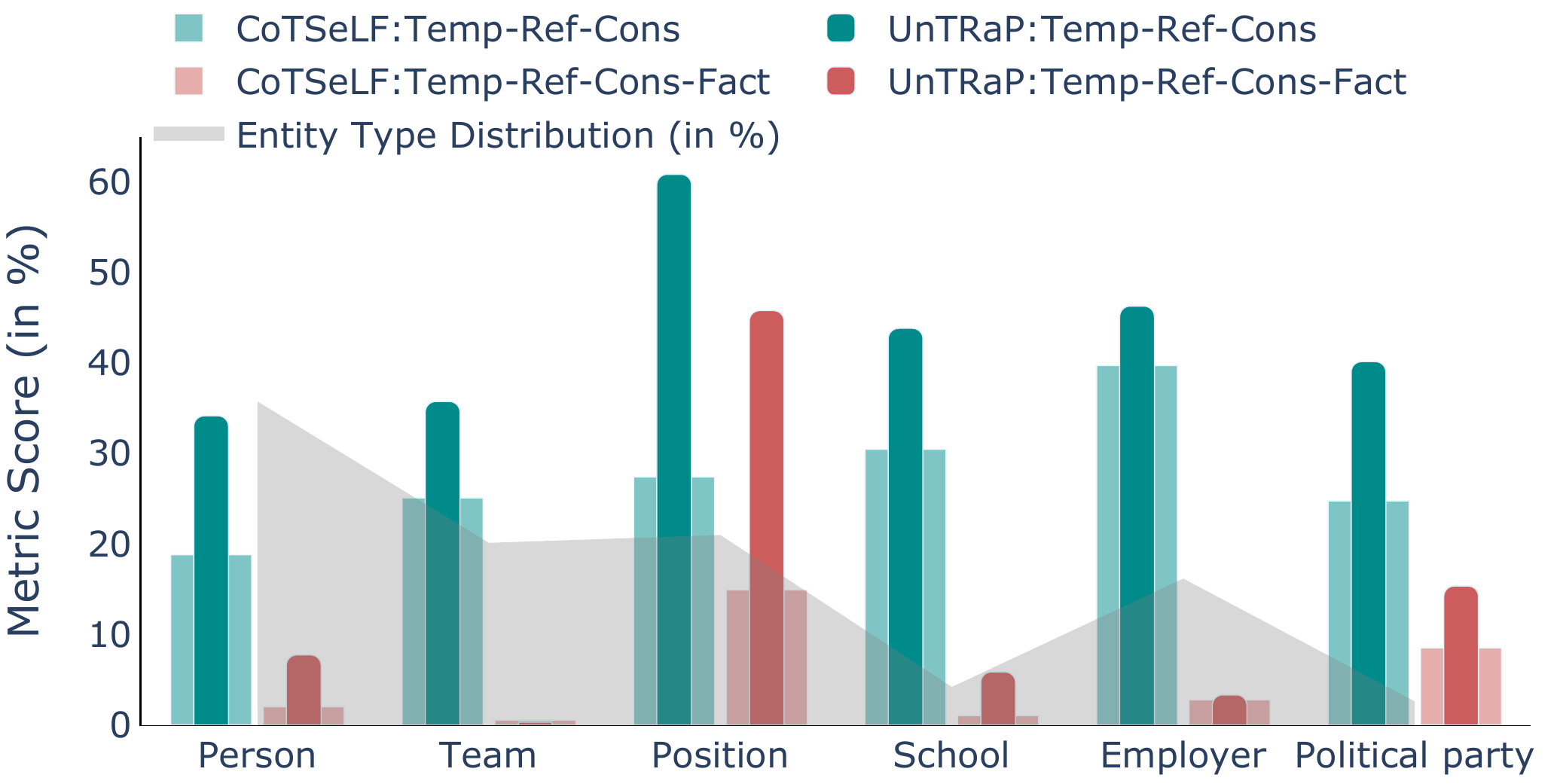}
  \caption{Temporally referential consistency across various entity types in the test data for \newmodel\ compared to the baseline CoTSeLF.
  }
  \label{fig:errorana}
  \vspace{-5mm}
\end{figure}
\paragraph{Scalability Across Base Models.} Table \ref{tab:res_gen} highlights the strong generalization capability of \newmodel\ across diverse LLMs, including both base and chat models with parameter sizes -- 8B and 13B. 
For LLaMA2-13B-Chat, temporal referential consistency scores improve to 43.81 (+30.71), demonstrating its ability to scale across different architectures and parameter configurations.

\paragraph{Unilateral Reasoning Path Fine-Tuning.} Here, we aim to assess the impact of unilateral reasoning path fine-tuning in comparison to \newmodel. We fine-tune the base model exclusively with instances derived from absolute temporal references, alongside the event-oriented reasoning path $r^a$. In Figure \ref{fig:unilateral}, a notable decline of 29.64 percentage points in temporal referential consistency is observed when contrasted with \newmodel, while no significant alteration in temporal referential factual deviation is detected. This finding is crucial and substantiates the argument for \newmodel, as widely employed unilateral reasoning path fine-tuning methods may substantially undermine the temporal referential consistency of the model.

\section{Error Analysis}

\begin{tcolorbox}[title=Qualitative Analysis, fonttitle=\small, boxrule=0.1mm, left=0mm, right=1mm]  
\small 
\textbf{Entity Type: Position}  
    \begin{itemize}  
        \item \textbf{Query}  
        \begin{itemize}  
            \item ABT: Which position did St John Brodrick, 1st Earl of Midleton hold right before November 1885?  
            \item CBT: Which position did St John Brodrick, 1st Earl of Midleton hold right before Member of the 23rd Parliament of the United Kingdom?  
        \end{itemize}  
        \item \textbf{Ground Truth Answer}: member of the 22nd parliament of the united kingdom  
        \item \textbf{CoTSeLF Response (Baseline)}   
        \begin{itemize}  
            \item ABT: member of the 23rd parliament of the united kingdom \textcolor{red}{X}  
            \item CBT: member of the 22nd parliament of the united kingdom \textcolor{green}{\checkmark}  
        \end{itemize}  
        \item \textbf{UnTRaP Response (Proposed Model)}   
        \begin{itemize}  
            \item ABT: member of the 22nd parliament of the united kingdom \textcolor{green}{\checkmark}  
            \item CBT: member of the 22nd parliament of the united kingdom \textcolor{green}{\checkmark}  
        \end{itemize}  
    \end{itemize}
\textbf{Entity Type: Team}  
    \begin{itemize}  
        \item \textbf{Query}  
        \begin{itemize}  
            \item ABT: Which team did Manuel Hartmann play for right after January 2010?  
            \item CBT: Which team did Manuel Hartmann play for right after TuS Koblenz?  
        \end{itemize}  
        \item \textbf{Ground Truth Answer}: FC Ingolstadt 04  
        \item \textbf{CoTSeLF Response (Baseline)}   
        \begin{itemize}  
            \item ABT: fc köln \textcolor{red}{X}  
            \item CBT: fc bayern munich \textcolor{red}{X}  
        \end{itemize}  
        \item \textbf{UnTRaP Response (Proposed Model)}   
        \begin{itemize}  
            \item ABT: fc ingolstadt 04 \textcolor{green}{\checkmark}  
            \item CBT: fc köln \textcolor{red}{X}  
        \end{itemize}  
    \end{itemize}  
\end{tcolorbox}
Here we evaluate the performance of the new model across various entity types to identify key error-contributing factors. In Figure \ref{fig:errorana}, we compare the temporal referential consistency and temporally referential consistent factuality among six entity types in the test data for \newmodel\ and the baseline, CoTSeLF. The correlation score between entity-type distribution and the new model's temporal referential consistency is -0.15, suggesting that entity sample size does not influence performance. In contrast, the correlation score between CoTSeLF and the proposed model's temporally referential consistent factuality is 0.95, indicating consistent factual performance distribution across entity types. These findings suggest that \newmodel\ \textbf{does not specifically target performance disparities among entity types within the foundation model}. Notably, the model shows disproportionate improvements of 33.44 and 30.86 percentage points for temporal referential consistency and consistent factuality for the `position' entity type. However, it fails to produce significant enhancements for the `team' name entity type, \textbf{highlighting the need for entity-specific adjustments within} \newmodel. An explicit example of such a case is presented in the qualitative analysis box, where \newmodel\ improves upon the baseline for the Position entity type but fails in the case of the Team entity type.

We also provide extended results in Appendix \ref{app:extended}, hyperparameters in Appendix \ref{app:hyper}, and commentaries on DeepSeek-R1, and domain adaptation and the applicability in closed-source models in Appendix \ref{app:extended} and Appendix \ref{App:commentary}, respectively.

\section{Related Work}
Research on temporal reasoning finds its roots in pioneering efforts such as TimeBank \cite{article} and TempEval \cite{verhagen-etal-2010-semeval}, laying the groundwork for understanding temporal expressions and their relationships in text, enabling more nuanced comprehension of temporal data. The evolution of Temporal Knowledge Graph Completion (TKGC) has led to the emergence of significant QA datasets, such as TEQUILA \cite{10.1145/3269206.3269247}, TimeQuestions \cite{10.1145/3459637.3482416}, and CronQuestions \cite{saxena-etal-2021-question}. Simultaneously, the consistency of LLMs \cite{liu2024trustworthyllmssurveyguideline} has been rigorously examined, yielding methods to assess inconsistencies across various tasks, including question answering \cite{kassner-etal-2020-pretrained,lee2024evaluatingconsistenciesllmresponses}, dialogue comprehension \cite{she-etal-2024-exploring,tang-etal-2022-confit,wang-etal-2022-analyzing}, summarization \cite{tam2023evaluatingfactualconsistencylarge,LUO2024124456,laban-etal-2023-summedits}, and natural language inference (NLI) \cite{li2019logicdrivenframeworkconsistencyneural}. Moreover, \citet{jain-etal-2023-language-models} conducted benchmarking of several LLMs using existing time-sensitive datasets.

The rapid deployment of language models in the public domain accentuates the imperative for both temporal accuracy and consistency within generated responses. Given this necessity, several time-sensitive QA datasets, such as TEMPLAMA \cite{dhingra-etal-2022-time} and TEMPREASON \cite{tan-etal-2023-towards}, have been established to evaluate and benchmark the temporal reasoning capabilities inherent in LLMs. Another pivotal contribution, TGQA (Temporal Graph QA) \cite{xiong2024largelanguagemodelslearn}, was proposed as a framework for the text-to-Temporal Graph translation task. Notably, TEMPREASON serves as a comprehensive benchmark for temporal reasoning, covering diverse temporal periods and three levels of temporal relations. \citet{bajpai-etal-2024-temporally} introduced the TEMP-COFAC dataset, revealing significant temporal consistency deficits in medium-sized LLMs. While \citet{kougia2024analysingzeroshottemporalrelation} assessed temporal consistency for relation extraction in biomedical text, their approach primarily focuses on surface changes in time-sensitive queries, neglecting consistency across temporal references. In contrast, our dataset, \newdata, addresses this critical measure of temporal consistency.

\section{Conclusion}
This paper presented a novel benchmark -- temporal referential consistency, along with a dataset \newdata\, designed to assess the sound temporal reasoning capabilities of LLMs. Our analysis revealed that all examined LLMs, across various linguistic contexts, demonstrate inadequate temporal referential consistency. 
Furthermore, we proposed \newmodel\, which unifies time- and event-oriented reasoning pathways across temporal references to enhance the model's temporal referential consistency. Empirical findings indicated that \newmodel\ outperforms several baseline models. The contributions and findings aim to augment public trust in LLMs, particularly in critical domains such as law and healthcare, thereby facilitating the adoption of LLMs for time-sensitive information processing.

\section{Limitations}
The proposed dataset, referred to as \newdata\, is designed for the evaluation of temporal referential consistency in large language models (LLMs). It is important to note that \newdata\ is derived from an existing dataset, TEMPREASON, and as such, we are reliant on the quality measures established by the original authors. Temporal referential consistency constitutes a specific aspect of temporal consistency (across temporal references); consequently, it necessitates integration with other prominent methodologies to effectively measure temporal consistency and to derive comprehensive metrics for LLM evaluation. 

Furthermore, recent literature has highlighted certain limitations inherent in the source dataset, such as its anchoring on a singular entity type, namely, person names, and a constrained range of entity types, which \newdata\ does not rectify. The proposed model, \newmodel\, employs a fine-tuning-based approach that necessitates access to the parameters of the foundational model. Consequently, the use of closed-source LLMs may be infeasible, as any benefits realized by \newmodel\, could inadvertently extend to these models. Additionally, the persistent challenges associated with computational resources for stacking LLMs further constrain our ability to experiment with models of larger parameter sizes.

\section{Ethical Considerations}
The dataset sourced in this study, referred to as TEMPREASON, is derived from the Wikidata Knowledge Base and Wikipedia articles, thereby adhering to the respective licensing agreements of these platforms. Specifically, Wikidata operates under a public domain\footnote{\url{https://www.wikidata.org/wiki/Wikidata:Licensing}} dedication, while Wikipedia articles are licensed under the Creative Commons Attribution-ShareAlike 3.0 Unported License\footnote{\url{https://en.wikipedia.org/wiki/Wikipedia:Copyrights}}. Consequently, our dataset will be disseminated under the same license as Wikidata, in accordance with the intentions of the original authors of TEMPREASON. 

It is important to delineate that the scope of the \newdata\ is strictly confined to the scientific investigation of temporal referential consistency within language models. Although TEMPREASON may encompass information pertaining to entities and their interrelations that are publicly accessible, any biases inherent in the selection of entities or the authenticity of content of the source data do not necessarily represent the perspectives or positions of the authors of this paper. In this study, all data annotators were compensated fairly for their contributions. The lack of women’s representation in \newdata\ annotations reflects global gender disparity in research and innovation, though the authors do not share this perspective. We genuinely support gender and racial equality in these fields. Additionally, no generative AI tools were used to create this content, aside from those for spell checking and grammar correction.

\section*{Acknowledgments}
Tanmoy Chakraborty acknowledges the support of Google GCP Grant,  IBM-IITD AI Horizons network and Rajiv Khemani Young Faculty
Chair Professorship in Artificial Intelligence.

\bibliography{camera_ready}

\begin{thebibliography}{47}
\providecommand{\natexlab}[1]{#1}

\bibitem[{Bajpai and Chakraborty(2025)}]{Bajpai_Chakraborty_2025}
Ashutosh Bajpai and Tanmoy Chakraborty. 2025.
\newblock \href {https://doi.org/10.1609/aaai.v39i22.34515} {Multilingual llms inherently reward in-language time-sensitive semantic alignment for low-resource languages}.
\newblock \emph{Proceedings of the AAAI Conference on Artificial Intelligence}, 39(22):23469--23477.

\bibitem[{Bajpai et~al.(2024)Bajpai, Goyal, Anwer, and Chakraborty}]{bajpai-etal-2024-temporally}
Ashutosh Bajpai, Aaryan Goyal, Atif Anwer, and Tanmoy Chakraborty. 2024.
\newblock \href {https://doi.org/10.18653/v1/2024.emnlp-main.887} {Temporally consistent factuality probing for large language models}.
\newblock In \emph{Proceedings of the 2024 Conference on Empirical Methods in Natural Language Processing}, pages 15864--15881, Miami, Florida, USA. Association for Computational Linguistics.

\bibitem[{Brown et~al.(2020)Brown, Mann, Ryder, Subbiah, Kaplan, Dhariwal, Neelakantan, Shyam, Sastry, Askell, Agarwal, Herbert-Voss, Krueger, Henighan, Child, Ramesh, Ziegler, Wu, Winter, Hesse, Chen, Sigler, Litwin, Gray, Chess, Clark, Berner, McCandlish, Radford, Sutskever, and Amodei}]{NEURIPS2020_1457c0d6}
Tom Brown, Benjamin Mann, Nick Ryder, Melanie Subbiah, Jared~D Kaplan, Prafulla Dhariwal, Arvind Neelakantan, Pranav Shyam, Girish Sastry, Amanda Askell, Sandhini Agarwal, Ariel Herbert-Voss, Gretchen Krueger, Tom Henighan, Rewon Child, Aditya Ramesh, Daniel Ziegler, Jeffrey Wu, Clemens Winter, and 12 others. 2020.
\newblock \href {https://proceedings.neurips.cc/paper_files/paper/2020/file/1457c0d6bfcb4967418bfb8ac142f64a-Paper.pdf} {Language models are few-shot learners}.
\newblock In \emph{Advances in Neural Information Processing Systems}, volume~33, pages 1877--1901. Curran Associates, Inc.

\bibitem[{Chung et~al.(2022)Chung, Hou, Longpre, Zoph, Tay, Fedus, Li, Wang, Dehghani, Brahma, Webson, Gu, Dai, Suzgun, Chen, Chowdhery, Castro-Ros, Pellat, Robinson, Valter, Narang, Mishra, Yu, Zhao, Huang, Dai, Yu, Petrov, Chi, Dean, Devlin, Roberts, Zhou, Le, and Wei}]{chung2022scalinginstructionfinetunedlanguagemodels}
Hyung~Won Chung, Le~Hou, Shayne Longpre, Barret Zoph, Yi~Tay, William Fedus, Yunxuan Li, Xuezhi Wang, Mostafa Dehghani, Siddhartha Brahma, Albert Webson, Shixiang~Shane Gu, Zhuyun Dai, Mirac Suzgun, Xinyun Chen, Aakanksha Chowdhery, Alex Castro-Ros, Marie Pellat, Kevin Robinson, and 16 others. 2022.
\newblock \href {https://arxiv.org/abs/2210.11416} {Scaling instruction-finetuned language models}.
\newblock \emph{Preprint}, arXiv:2210.11416.

\bibitem[{Dhingra et~al.(2022)Dhingra, Cole, Eisenschlos, Gillick, Eisenstein, and Cohen}]{dhingra-etal-2022-time}
Bhuwan Dhingra, Jeremy~R. Cole, Julian~Martin Eisenschlos, Daniel Gillick, Jacob Eisenstein, and William~W. Cohen. 2022.
\newblock \href {https://doi.org/10.1162/tacl_a_00459} {Time-aware language models as temporal knowledge bases}.
\newblock \emph{Transactions of the Association for Computational Linguistics}, 10:257--273.

\bibitem[{Elazar et~al.(2021)Elazar, Kassner, Ravfogel, Ravichander, Hovy, Sch{\"u}tze, and Goldberg}]{elazar-etal-2021-measuring}
Yanai Elazar, Nora Kassner, Shauli Ravfogel, Abhilasha Ravichander, Eduard Hovy, Hinrich Sch{\"u}tze, and Yoav Goldberg. 2021.
\newblock \href {https://doi.org/10.1162/tacl_a_00410} {Measuring and improving consistency in pretrained language models}.
\newblock \emph{Transactions of the Association for Computational Linguistics}, 9:1012--1031.

\bibitem[{Grattafiori et~al.(2024)Grattafiori, Dubey, and Others}]{grattafiori2024llama3herdmodels}
Aaron Grattafiori, Abhimanyu Dubey, and Others. 2024.
\newblock \href {https://arxiv.org/abs/2407.21783} {The llama 3 herd of models}.
\newblock \emph{Preprint}, arXiv:2407.21783.

\bibitem[{Hu et~al.(2021)Hu, Shen, Wallis, Allen-Zhu, Li, Wang, Wang, and Chen}]{hu2021loralowrankadaptationlarge}
Edward~J. Hu, Yelong Shen, Phillip Wallis, Zeyuan Allen-Zhu, Yuanzhi Li, Shean Wang, Lu~Wang, and Weizhu Chen. 2021.
\newblock \href {https://arxiv.org/abs/2106.09685} {Lora: Low-rank adaptation of large language models}.
\newblock \emph{Preprint}, arXiv:2106.09685.

\bibitem[{Jain et~al.(2023)Jain, Sojitra, Acharya, Saha, Jatowt, and Dandapat}]{jain-etal-2023-language-models}
Raghav Jain, Daivik Sojitra, Arkadeep Acharya, Sriparna Saha, Adam Jatowt, and Sandipan Dandapat. 2023.
\newblock \href {https://doi.org/10.18653/v1/2023.emnlp-main.418} {Do language models have a common sense regarding time? revisiting temporal commonsense reasoning in the era of large language models}.
\newblock In \emph{Proceedings of the 2023 Conference on Empirical Methods in Natural Language Processing}, pages 6750--6774, Singapore. Association for Computational Linguistics.

\bibitem[{Jia et~al.(2018)Jia, Abujabal, Saha~Roy, Str\"{o}tgen, and Weikum}]{10.1145/3269206.3269247}
Zhen Jia, Abdalghani Abujabal, Rishiraj Saha~Roy, Jannik Str\"{o}tgen, and Gerhard Weikum. 2018.
\newblock \href {https://doi.org/10.1145/3269206.3269247} {Tequila: Temporal question answering over knowledge bases}.
\newblock In \emph{Proceedings of the 27th ACM International Conference on Information and Knowledge Management}, CIKM '18, page 1807–1810, New York, NY, USA. Association for Computing Machinery.

\bibitem[{Jia et~al.(2021)Jia, Pramanik, Saha~Roy, and Weikum}]{10.1145/3459637.3482416}
Zhen Jia, Soumajit Pramanik, Rishiraj Saha~Roy, and Gerhard Weikum. 2021.
\newblock \href {https://doi.org/10.1145/3459637.3482416} {Complex temporal question answering on knowledge graphs}.
\newblock In \emph{Proceedings of the 30th ACM International Conference on Information \& Knowledge Management}, CIKM '21, page 792–802, New York, NY, USA. Association for Computing Machinery.

\bibitem[{Jiang et~al.(2023)Jiang, Sablayrolles, Mensch, Bamford, Chaplot, de~las Casas, Bressand, Lengyel, Lample, Saulnier, Lavaud, Lachaux, Stock, Scao, Lavril, Wang, Lacroix, and Sayed}]{jiang2023mistral7b}
Albert~Q. Jiang, Alexandre Sablayrolles, Arthur Mensch, Chris Bamford, Devendra~Singh Chaplot, Diego de~las Casas, Florian Bressand, Gianna Lengyel, Guillaume Lample, Lucile Saulnier, Lélio~Renard Lavaud, Marie-Anne Lachaux, Pierre Stock, Teven~Le Scao, Thibaut Lavril, Thomas Wang, Timothée Lacroix, and William~El Sayed. 2023.
\newblock \href {https://arxiv.org/abs/2310.06825} {Mistral 7b}.
\newblock \emph{Preprint}, arXiv:2310.06825.

\bibitem[{Kassner et~al.(2020)Kassner, Krojer, and Sch{\"u}tze}]{kassner-etal-2020-pretrained}
Nora Kassner, Benno Krojer, and Hinrich Sch{\"u}tze. 2020.
\newblock \href {https://doi.org/10.18653/v1/2020.conll-1.45} {Are pretrained language models symbolic reasoners over knowledge?}
\newblock In \emph{Proceedings of the 24th Conference on Computational Natural Language Learning}, pages 552--564, Online. Association for Computational Linguistics.

\bibitem[{Kougia et~al.(2024)Kougia, Sedova, Stephan, Zaporojets, and Roth}]{kougia2024analysingzeroshottemporalrelation}
Vasiliki Kougia, Anastasiia Sedova, Andreas Stephan, Klim Zaporojets, and Benjamin Roth. 2024.
\newblock \href {https://arxiv.org/abs/2406.11486} {Analysing zero-shot temporal relation extraction on clinical notes using temporal consistency}.
\newblock \emph{Preprint}, arXiv:2406.11486.

\bibitem[{Laban et~al.(2023)Laban, Kryscinski, Agarwal, Fabbri, Xiong, Joty, and Wu}]{laban-etal-2023-summedits}
Philippe Laban, Wojciech Kryscinski, Divyansh Agarwal, Alexander Fabbri, Caiming Xiong, Shafiq Joty, and Chien-Sheng Wu. 2023.
\newblock \href {https://doi.org/10.18653/v1/2023.emnlp-main.600} {{S}umm{E}dits: Measuring {LLM} ability at factual reasoning through the lens of summarization}.
\newblock In \emph{Proceedings of the 2023 Conference on Empirical Methods in Natural Language Processing}, pages 9662--9676, Singapore. Association for Computational Linguistics.

\bibitem[{Lee and Kim(2024)}]{lee2024evaluatingconsistenciesllmresponses}
Yanggyu Lee and Jihie Kim. 2024.
\newblock \href {https://arxiv.org/abs/2410.15440} {Evaluating consistencies in llm responses through a semantic clustering of question answering}.
\newblock \emph{Preprint}, arXiv:2410.15440.

\bibitem[{Li et~al.(2019)Li, Gupta, Mehta, and Srikumar}]{li2019logicdrivenframeworkconsistencyneural}
Tao Li, Vivek Gupta, Maitrey Mehta, and Vivek Srikumar. 2019.
\newblock \href {https://arxiv.org/abs/1909.00126} {A logic-driven framework for consistency of neural models}.
\newblock \emph{Preprint}, arXiv:1909.00126.

\bibitem[{Liu et~al.(2022)Liu, Shen, Zhang, Dolan, Carin, and Chen}]{liu-etal-2022-makes}
Jiachang Liu, Dinghan Shen, Yizhe Zhang, Bill Dolan, Lawrence Carin, and Weizhu Chen. 2022.
\newblock \href {https://doi.org/10.18653/v1/2022.deelio-1.10} {What makes good in-context examples for {GPT}-3?}
\newblock In \emph{Proceedings of Deep Learning Inside Out (DeeLIO 2022): The 3rd Workshop on Knowledge Extraction and Integration for Deep Learning Architectures}, pages 100--114, Dublin, Ireland and Online. Association for Computational Linguistics.

\bibitem[{Liu et~al.(2024)Liu, Yao, Ton, Zhang, Guo, Cheng, Klochkov, Taufiq, and Li}]{liu2024trustworthyllmssurveyguideline}
Yang Liu, Yuanshun Yao, Jean-Francois Ton, Xiaoying Zhang, Ruocheng Guo, Hao Cheng, Yegor Klochkov, Muhammad~Faaiz Taufiq, and Hang Li. 2024.
\newblock \href {https://arxiv.org/abs/2308.05374} {Trustworthy llms: a survey and guideline for evaluating large language models' alignment}.
\newblock \emph{Preprint}, arXiv:2308.05374.

\bibitem[{Lu and Naseem(2024)}]{lu-naseem-2024-large}
Haohui Lu and Usman Naseem. 2024.
\newblock \href {https://doi.org/10.18653/v1/2024.emnlp-main.980} {Can large language models enhance predictions of disease progression? investigating through disease network link prediction}.
\newblock In \emph{Proceedings of the 2024 Conference on Empirical Methods in Natural Language Processing}, pages 17703--17715, Miami, Florida, USA. Association for Computational Linguistics.

\bibitem[{Luo et~al.(2024)Luo, Xie, and Ananiadou}]{LUO2024124456}
Zheheng Luo, Qianqian Xie, and Sophia Ananiadou. 2024.
\newblock \href {https://doi.org/10.1016/j.eswa.2024.124456} {Factual consistency evaluation of summarization in the era of large language models}.
\newblock \emph{Expert Systems with Applications}, 254:124456.

\bibitem[{Muennighoff et~al.(2023)Muennighoff, Wang, Sutawika, Roberts, Biderman, Le~Scao, Bari, Shen, Yong, Schoelkopf, Tang, Radev, Aji, Almubarak, Albanie, Alyafeai, Webson, Raff, and Raffel}]{muennighoff-etal-2023-crosslingual}
Niklas Muennighoff, Thomas Wang, Lintang Sutawika, Adam Roberts, Stella Biderman, Teven Le~Scao, M~Saiful Bari, Sheng Shen, Zheng~Xin Yong, Hailey Schoelkopf, Xiangru Tang, Dragomir Radev, Alham~Fikri Aji, Khalid Almubarak, Samuel Albanie, Zaid Alyafeai, Albert Webson, Edward Raff, and Colin Raffel. 2023.
\newblock \href {https://doi.org/10.18653/v1/2023.acl-long.891} {Crosslingual generalization through multitask finetuning}.
\newblock In \emph{Proceedings of the 61st Annual Meeting of the Association for Computational Linguistics (Volume 1: Long Papers)}, pages 15991--16111, Toronto, Canada. Association for Computational Linguistics.

\bibitem[{Narendra et~al.(2024)Narendra, Shetty, and Ratnaparkhi}]{narendra-etal-2024-enhancing}
Savinay Narendra, Kaushal Shetty, and Adwait Ratnaparkhi. 2024.
\newblock \href {https://doi.org/10.18653/v1/2024.nllp-1.11} {Enhancing contract negotiations with {LLM}-based legal document comparison}.
\newblock In \emph{Proceedings of the Natural Legal Language Processing Workshop 2024}, pages 143--153, Miami, FL, USA. Association for Computational Linguistics.

\bibitem[{OpenAI et~al.(2024)OpenAI, Achiam, and Others}]{openai2024gpt4technicalreport}
OpenAI, Josh Achiam, and Others. 2024.
\newblock \href {https://arxiv.org/abs/2303.08774} {Gpt-4 technical report}.
\newblock \emph{Preprint}, arXiv:2303.08774.

\bibitem[{Ouyang et~al.(2022)Ouyang, Wu, Jiang, Almeida, Wainwright, Mishkin, Zhang, Agarwal, Slama, Ray, Schulman, Hilton, Kelton, Miller, Simens, Askell, Welinder, Christiano, Leike, and Lowe}]{10.5555/3600270.3602281}
Long Ouyang, Jeff Wu, Xu~Jiang, Diogo Almeida, Carroll~L. Wainwright, Pamela Mishkin, Chong Zhang, Sandhini Agarwal, Katarina Slama, Alex Ray, John Schulman, Jacob Hilton, Fraser Kelton, Luke Miller, Maddie Simens, Amanda Askell, Peter Welinder, Paul Christiano, Jan Leike, and Ryan Lowe. 2022.
\newblock Training language models to follow instructions with human feedback.
\newblock In \emph{Proceedings of the 36th International Conference on Neural Information Processing Systems}, NIPS '22, Red Hook, NY, USA. Curran Associates Inc.

\bibitem[{Petroni et~al.(2019)Petroni, Rocktäschel, Lewis, Bakhtin, Wu, Miller, and Riedel}]{petroni2019languagemodelsknowledgebases}
Fabio Petroni, Tim Rocktäschel, Patrick Lewis, Anton Bakhtin, Yuxiang Wu, Alexander~H. Miller, and Sebastian Riedel. 2019.
\newblock \href {https://arxiv.org/abs/1909.01066} {Language models as knowledge bases?}
\newblock \emph{Preprint}, arXiv:1909.01066.

\bibitem[{Popovi{\'c}(2015)}]{popovic-2015-chrf}
Maja Popovi{\'c}. 2015.
\newblock \href {https://doi.org/10.18653/v1/W15-3049} {chr{F}: character n-gram {F}-score for automatic {MT} evaluation}.
\newblock In \emph{Proceedings of the Tenth Workshop on Statistical Machine Translation}, pages 392--395, Lisbon, Portugal. Association for Computational Linguistics.

\bibitem[{Pustejovsky et~al.(2003)Pustejovsky, Hanks, Saurí, See, Gaizauskas, Setzer, Radev, Sundheim, Day, Ferro, and Lazo}]{article}
James Pustejovsky, Patrick Hanks, Roser Saurí, Andrew See, Rob Gaizauskas, Andrea Setzer, Dragomir Radev, Beth Sundheim, David Day, Lisa Ferro, and Marcia Lazo. 2003.
\newblock The timebank corpus.
\newblock \emph{Proceedings of Corpus Linguistics}.

\bibitem[{Raffel et~al.(2023)Raffel, Shazeer, Roberts, Lee, Narang, Matena, Zhou, Li, and Liu}]{raffel2023exploringlimitstransferlearning}
Colin Raffel, Noam Shazeer, Adam Roberts, Katherine Lee, Sharan Narang, Michael Matena, Yanqi Zhou, Wei Li, and Peter~J. Liu. 2023.
\newblock \href {https://arxiv.org/abs/1910.10683} {Exploring the limits of transfer learning with a unified text-to-text transformer}.
\newblock \emph{Preprint}, arXiv:1910.10683.

\bibitem[{Ravichander et~al.(2020)Ravichander, Hovy, Suleman, Trischler, and Cheung}]{ravichander-etal-2020-systematicity}
Abhilasha Ravichander, Eduard Hovy, Kaheer Suleman, Adam Trischler, and Jackie Chi~Kit Cheung. 2020.
\newblock \href {https://aclanthology.org/2020.starsem-1.10/} {On the systematicity of probing contextualized word representations: The case of hypernymy in {BERT}}.
\newblock In \emph{Proceedings of the Ninth Joint Conference on Lexical and Computational Semantics}, pages 88--102, Barcelona, Spain (Online). Association for Computational Linguistics.

\bibitem[{Saxena et~al.(2021)Saxena, Chakrabarti, and Talukdar}]{saxena-etal-2021-question}
Apoorv Saxena, Soumen Chakrabarti, and Partha Talukdar. 2021.
\newblock \href {https://doi.org/10.18653/v1/2021.acl-long.520} {Question answering over temporal knowledge graphs}.
\newblock In \emph{Proceedings of the 59th Annual Meeting of the Association for Computational Linguistics and the 11th International Joint Conference on Natural Language Processing (Volume 1: Long Papers)}, pages 6663--6676, Online. Association for Computational Linguistics.

\bibitem[{She et~al.(2024)She, Huang, Wang, Zhou, and Chen}]{she-etal-2024-exploring}
Shuaijie She, Shujian Huang, Xingyun Wang, Yanke Zhou, and Jiajun Chen. 2024.
\newblock \href {https://doi.org/10.18653/v1/2024.naacl-long.338} {Exploring the factual consistency in dialogue comprehension of large language models}.
\newblock In \emph{Proceedings of the 2024 Conference of the North American Chapter of the Association for Computational Linguistics: Human Language Technologies (Volume 1: Long Papers)}, pages 6087--6100, Mexico City, Mexico. Association for Computational Linguistics.

\bibitem[{Sojitra et~al.(2024)Sojitra, Jain, Saha, Jatowt, and Gupta}]{10.1145/3626772.3657899}
Daivik Sojitra, Raghav Jain, Sriparna Saha, Adam Jatowt, and Manish Gupta. 2024.
\newblock \href {https://doi.org/10.1145/3626772.3657899} {Timeline summarization in the era of llms}.
\newblock In \emph{Proceedings of the 47th International ACM SIGIR Conference on Research and Development in Information Retrieval}, SIGIR '24, page 2657–2661, New York, NY, USA. Association for Computing Machinery.

\bibitem[{Tam et~al.(2023{\natexlab{a}})Tam, Mascarenhas, Zhang, Kwan, Bansal, and Raffel}]{tam-etal-2023-evaluating}
Derek Tam, Anisha Mascarenhas, Shiyue Zhang, Sarah Kwan, Mohit Bansal, and Colin Raffel. 2023{\natexlab{a}}.
\newblock \href {https://doi.org/10.18653/v1/2023.findings-acl.322} {Evaluating the factual consistency of large language models through news summarization}.
\newblock In \emph{Findings of the Association for Computational Linguistics: ACL 2023}, pages 5220--5255, Toronto, Canada. Association for Computational Linguistics.

\bibitem[{Tam et~al.(2023{\natexlab{b}})Tam, Mascarenhas, Zhang, Kwan, Bansal, and Raffel}]{tam2023evaluatingfactualconsistencylarge}
Derek Tam, Anisha Mascarenhas, Shiyue Zhang, Sarah Kwan, Mohit Bansal, and Colin Raffel. 2023{\natexlab{b}}.
\newblock \href {https://arxiv.org/abs/2211.08412} {Evaluating the factual consistency of large language models through news summarization}.
\newblock \emph{Preprint}, arXiv:2211.08412.

\bibitem[{Tan et~al.(2023)Tan, Ng, and Bing}]{tan-etal-2023-towards}
Qingyu Tan, Hwee~Tou Ng, and Lidong Bing. 2023.
\newblock \href {https://doi.org/10.18653/v1/2023.acl-long.828} {Towards benchmarking and improving the temporal reasoning capability of large language models}.
\newblock In \emph{Proceedings of the 61st Annual Meeting of the Association for Computational Linguistics (Volume 1: Long Papers)}, pages 14820--14835, Toronto, Canada. Association for Computational Linguistics.

\bibitem[{Tang et~al.(2022)Tang, Nair, Wang, Wang, Desai, Wade, Li, Celikyilmaz, Mehdad, and Radev}]{tang-etal-2022-confit}
Xiangru Tang, Arjun Nair, Borui Wang, Bingyao Wang, Jai Desai, Aaron Wade, Haoran Li, Asli Celikyilmaz, Yashar Mehdad, and Dragomir Radev. 2022.
\newblock \href {https://doi.org/10.18653/v1/2022.naacl-main.415} {{CONFIT}: Toward faithful dialogue summarization with linguistically-informed contrastive fine-tuning}.
\newblock In \emph{Proceedings of the 2022 Conference of the North American Chapter of the Association for Computational Linguistics: Human Language Technologies}, pages 5657--5668, Seattle, United States. Association for Computational Linguistics.

\bibitem[{Taori et~al.(2023)Taori, Gulrajani, Zhang, Dubois, Li, Guestrin, Liang, and Hashimoto}]{alpaca}
Rohan Taori, Ishaan Gulrajani, Tianyi Zhang, Yann Dubois, Xuechen Li, Carlos Guestrin, Percy Liang, and Tatsunori~B. Hashimoto. 2023.
\newblock Stanford alpaca: An instruction-following llama model.
\newblock \url{https://github.com/tatsu-lab/stanford_alpaca}.

\bibitem[{Team et~al.(2024)Team, Georgiev, and Others}]{geminiteam2024gemini15unlockingmultimodal}
Gemini Team, Petko Georgiev, and Others. 2024.
\newblock \href {https://arxiv.org/abs/2403.05530} {Gemini 1.5: Unlocking multimodal understanding across millions of tokens of context}.
\newblock \emph{Preprint}, arXiv:2403.05530.

\bibitem[{Team et~al.(2022)Team, Costa-jussà, Cross, Çelebi, Elbayad, Heafield, Heffernan, Kalbassi, Lam, Licht, Maillard, Sun, Wang, Wenzek, Youngblood, Akula, Barrault, Gonzalez, Hansanti, Hoffman, Jarrett, Sadagopan, Rowe, Spruit, Tran, Andrews, Ayan, Bhosale, Edunov, Fan, Gao, Goswami, Guzmán, Koehn, Mourachko, Ropers, Saleem, Schwenk, and Wang}]{nllbteam2022languageleftbehindscaling}
NLLB Team, Marta~R. Costa-jussà, James Cross, Onur Çelebi, Maha Elbayad, Kenneth Heafield, Kevin Heffernan, Elahe Kalbassi, Janice Lam, Daniel Licht, Jean Maillard, Anna Sun, Skyler Wang, Guillaume Wenzek, Al~Youngblood, Bapi Akula, Loic Barrault, Gabriel~Mejia Gonzalez, Prangthip Hansanti, and 20 others. 2022.
\newblock \href {https://arxiv.org/abs/2207.04672} {No language left behind: Scaling human-centered machine translation}.
\newblock \emph{Preprint}, arXiv:2207.04672.

\bibitem[{Verhagen et~al.(2010)Verhagen, Saur{\'i}, Caselli, and Pustejovsky}]{verhagen-etal-2010-semeval}
Marc Verhagen, Roser Saur{\'i}, Tommaso Caselli, and James Pustejovsky. 2010.
\newblock \href {https://aclanthology.org/S10-1010/} {{S}em{E}val-2010 task 13: {T}emp{E}val-2}.
\newblock In \emph{Proceedings of the 5th International Workshop on Semantic Evaluation}, pages 57--62, Uppsala, Sweden. Association for Computational Linguistics.

\bibitem[{Wang et~al.(2022)Wang, Zhang, Zhang, Chen, and Li}]{wang-etal-2022-analyzing}
Bin Wang, Chen Zhang, Yan Zhang, Yiming Chen, and Haizhou Li. 2022.
\newblock \href {https://doi.org/10.18653/v1/2022.emnlp-main.325} {Analyzing and evaluating faithfulness in dialogue summarization}.
\newblock In \emph{Proceedings of the 2022 Conference on Empirical Methods in Natural Language Processing}, pages 4897--4908, Abu Dhabi, United Arab Emirates. Association for Computational Linguistics.

\bibitem[{Wei et~al.(2022)Wei, Wang, Schuurmans, Bosma, ichter, Xia, Chi, Le, and Zhou}]{NEURIPS2022_9d560961}
Jason Wei, Xuezhi Wang, Dale Schuurmans, Maarten Bosma, brian ichter, Fei Xia, Ed~Chi, Quoc~V Le, and Denny Zhou. 2022.
\newblock \href {https://proceedings.neurips.cc/paper_files/paper/2022/file/9d5609613524ecf4f15af0f7b31abca4-Paper-Conference.pdf} {Chain-of-thought prompting elicits reasoning in large language models}.
\newblock In \emph{Advances in Neural Information Processing Systems}, volume~35, pages 24824--24837. Curran Associates, Inc.

\bibitem[{Xiong et~al.(2024)Xiong, Payani, Kompella, and Fekri}]{xiong2024largelanguagemodelslearn}
Siheng Xiong, Ali Payani, Ramana Kompella, and Faramarz Fekri. 2024.
\newblock \href {https://arxiv.org/abs/2401.06853} {Large language models can learn temporal reasoning}.
\newblock \emph{Preprint}, arXiv:2401.06853.

\bibitem[{Yuan et~al.(2023)Yuan, Yuan, Li, Dong, Lu, Tan, Zhou, and Zhou}]{yuan2023scalingrelationshiplearningmathematical}
Zheng Yuan, Hongyi Yuan, Chengpeng Li, Guanting Dong, Keming Lu, Chuanqi Tan, Chang Zhou, and Jingren Zhou. 2023.
\newblock \href {https://arxiv.org/abs/2308.01825} {Scaling relationship on learning mathematical reasoning with large language models}.
\newblock \emph{Preprint}, arXiv:2308.01825.

\bibitem[{Zeng(2024)}]{zeng2024histolensllmpoweredframeworkmultilayered}
Yifan Zeng. 2024.
\newblock \href {https://arxiv.org/abs/2411.09978} {Histolens: An llm-powered framework for multi-layered analysis of historical texts -- a case application of yantie lun}.
\newblock \emph{Preprint}, arXiv:2411.09978.

\bibitem[{Zheng et~al.(2023)Zheng, Chiang, Sheng, Zhuang, Wu, Zhuang, Lin, Li, Li, Xing, Zhang, Gonzalez, and Stoica}]{zheng2023judgingllmasajudgemtbenchchatbot}
Lianmin Zheng, Wei-Lin Chiang, Ying Sheng, Siyuan Zhuang, Zhanghao Wu, Yonghao Zhuang, Zi~Lin, Zhuohan Li, Dacheng Li, Eric~P. Xing, Hao Zhang, Joseph~E. Gonzalez, and Ion Stoica. 2023.
\newblock \href {https://arxiv.org/abs/2306.05685} {Judging llm-as-a-judge with mt-bench and chatbot arena}.
\newblock \emph{Preprint}, arXiv:2306.05685.

\end{thebibliography}
\appendix
\label{sec:appendix}

\section{Extended Results}
\label{app:extended}
\paragraph{Experimental Settings.} We primarily utilize LLaMA3.1-8B \cite{grattafiori2024llama3herdmodels} for all experimental procedures, including implementing our \newmodel\ model. We opt for supervised instruction-tuning (IT) \cite{10.5555/3600270.3602281} to enhance the temporally referential consistency of the pre-trained model. This approach aligns with parameter-efficient fine-tuning (PEFT) techniques, such as low-rank adaptation (LoRA) \cite{hu2021loralowrankadaptationlarge}, which provide substantial advantages for instruction tuning within cost-effective infrastructures. By leveraging advancements in model fine-tuning alongside the train set of \newdata, we aim to fine-tune a task-specific adapter. We utilize a test set to benchmark all the results presented in this study. Our experimental framework uses a three-shot methodology for non-gradient update baselines such as in-context learning (ICL) and chain-of-thought (CoT) reasoning. Further, we use EM-based temporal factuality measure to calculate temporal referential consistency and temporally referential consistent factuality.
\paragraph{Baselines.} Here is the details on all the baselines used in this work -
\begin{itemize}
[noitemsep,nolistsep,topsep=0pt,leftmargin=1em]
    \item \textbf{In-Context Learning} (ICL) \cite{NEURIPS2020_1457c0d6}: The model is primed with a set of instances from training data, which serve as demonstrations, accompanied by a query. 
    \item \textbf{Semantic In-Context Learning} (Semantic ICL) \cite{liu-etal-2022-makes}: The Semantic-ICL approach enhances this process by incorporating the retrieval of semantically similar examples for queries from training data.
    \item \textbf{Semantic Chain-of-Thoughts Reasoning} (Semantic CoT): We augment the standard CoT \cite{NEURIPS2022_9d560961} approach by drawing semantically aligned examples and respective reasoning steps for this baseline.
    \item \textbf{SFT Model} (Instruction Tuning \cite{10.5555/3600270.3602281}: The model is fine-tuned on a supervised dataset of instruction, input and corresponding responses, helping it better understand and follow specific instructions.
    \item \textbf{SFT + Time-Sensitive Reinforcement Learning} (TSRL) \cite{tan-etal-2023-towards}: TSRL positively reinforces the model for identifying the correct temporal response among various temporal possibilities. In this approach, we utilize a closed-vocabulary framework, restricting the model to its factual knowledge only.
    \item \textbf{CoTSeLF: MT-IT + Consistent Time-Sensitive Reinforcement Learning} \cite{bajpai-etal-2024-temporally}: An explicit binary task to enhance temporal consistency is fine-tuned in conjunction with the primary objective of ensuring temporal factuality followed by a reward-based framework to reinforce the model.
\end{itemize}
\paragraph{Unilateral Reasoning Path-based Fine-Tuning.} The detailed results for the temporal referential consistency for the \newmodel\ compared to unilateral reasoning path based fine-tuning is presented in Table \ref{tab:unilateral}.

\begin{table*}[!t]
\centering
\small
\resizebox{.78\textwidth}{!}{
\begin{tabular}{l|c|c|c|c}
\toprule
\textbf{Method}	&\multicolumn{2}{c|}{\textbf{Temp-Ref-Fact-Dev}}&	\textbf{Temp-Ref-Cons}&	\textbf{Temp-Ref-Cons-Fact}\\
\cline{2-3}
& \textbf{\hspace{2.7mm}EM\hspace{2.7mm}} &\textbf{F1.} & & \\
\midrule
\newmodel	&-0.45&	-0.01	&42.63	&13.67\\
Unilateral Fine-tuning	&-1.61	&-2.22	&12.99	&7.14\\
\bottomrule
\end{tabular}
}
\caption{Detailed Results: a comparison between \newmodel\ with unilateral (event-oriented) reasoning path fine-tuning utilizing only absolute temporal reference-based training instances.}
\label{tab:unilateral}
\end{table*}
\paragraph{Extended Analysis of Commercial LLMs.} Additionally, in our extended evaluation of closed-source LLMs such as GPT-4 \cite{openai2024gpt4technicalreport} and Claude-3-Opus\footnote{\url{https://www.anthropic.com/news/claude-3-family}}, we observe that they also demonstrate suboptimal performance regarding temporal references, presented in Table \ref{tab:app_commercial_llm}. The temporal referential factuality deviation for GPT is quantified at -6.40 and -4.69 for EM and F1-based metrics, respectively, representing the lowest performance among all LLMs assessed in this study, thereby indicating an urgent need for improvement. For benchmarking commercial LLMs, we randomly subsample the test data to a thousand instances due to limited access attributed to financial constraints.
\begin{table*}[t]
\centering
\small
\resizebox{.98\textwidth}{!}{
\begin{tabular}{l|l|c|c|c|c|c|c|c|c}
\toprule
\textbf{Model}	&\textbf{Method}	&\multicolumn{6}{c|}{\textbf{Temp-Ref-Fact}}&	\textbf{Temp-Ref-Cons}&	\textbf{Temp-Ref-Cons-Fact}\\
\cline{3-8}
& &\multicolumn{3}{c|}{\textbf{EM}} &\multicolumn{3}{c|}{\textbf{F1.}} & & \\
\cline{3-8}
& &\textbf{CTR}&\textbf{ATR}&\textbf{Dev.} &\textbf{CTR}&\textbf{ATR}&\textbf{Dev.} & & \\
\midrule
\multicolumn{10}{c}{\textit{Closed-Source LLMs}}\\
\midrule
GPT-4&Zero-Shot	&22.5&16.1&-6.40&31.89&27.20	&-4.69&	27.20&	9.60\\
Claude-3-Opus&Zero-Shot	&13.9&12.8&-1.10	&24.93&23.26&-1.67&	23.20&	8.00\\
\bottomrule
\end{tabular}
}
\caption{Zero-shot performance of Commercial LLMs on temporal referential consistency.}
\label{tab:app_commercial_llm}
\end{table*}

\paragraph{Error Analysis.} The detailed results for the temporal referential consistency across entity types is presented in Table \ref{tab:erroranafull}.

\begin{table*}[!t]
\centering
\small
\resizebox{.98\textwidth}{!}{
\begin{tabular}{l|l|c|c|c|c|c|c}
\toprule
\textbf{Model}	&\textbf{Metric} &\textbf{Person} &\textbf{Team} &\textbf{Position} &\textbf{School} &\textbf{Employer} &\textbf{Political party}\\
\midrule
& \#Count of Queries & 1583	&892&	930&	187&	717&	117\\
\midrule
\multirow{2}*{CoTSeLF} & Temp-Ref-Cons&18.83&	25.11	&27.42	&30.48&	39.75	&24.79	\\
& Temp-Ref-Cons-Fact&2.02&	0.56	&14.95	&1.07&	2.79&	8.55\\
\midrule
\multirow{2}*{\newmodel} & Temp-Ref-Cons&34.18	&35.76&	60.86	&43.85	&46.3&	40.17	\\
& Temp-Ref-Cons-Fact&7.77&	0.34	&45.81	&5.88&	3.35&	15.38\\

\bottomrule
\end{tabular}
}
\caption{Temporally referential consistency across various entity types in the test data for \newmodel\ compared to the baseline CoTSeLF along with the count of queries in test set for respective entity types.}
\label{tab:erroranafull}
\end{table*}

\paragraph{DeepSeek-R1 Qualitative Analysis.} DeepSeek-R1 is a flagship model, showcasing strong reasoning capabilities across domain and tasks. We put the DeepSeek-R1 on test for our scenario. We test the model qualitatively across temporal directions before/after and temporal references- absolute and chronological. The reasoning capabilities of DeepSeek-R1, as demonstrated in its responses, presented in Table \ref{tab:deepseek}, to questions about the head coach history of Legia Warsaw, showcase several significant challenges with accuracy, consistency, and contextual understanding:

\begin{itemize}
[itemsep=2pt,nolistsep,topsep=0pt,leftmargin=1em]
    \item \textbf{Inconsistent Temporal Reasoning:} DeepSeek-R1 fails to align its answers with the factual timeline provided. For example, in response to the first question, it incorrectly states that Edmund Zientara left Legia Warsaw in 1980, when the actual timeline indicates he was head coach from July 1969 to July 1971. Similarly, Józef Zwierzyna, mentioned as succeeding Zientara in 1980, does not appear in the factual data provided, showing a lack of alignment with the given information.
    \item \textbf{Circular Errors in Temporal Placement:} The second question's response claims that Zientara became head coach in July 1971, which contradicts the provided timeline, where Tadeusz Chruściński was listed as the coach after Zientara post-July 1971.
    \item \textbf{Fabrication of Unverified Data:} In the third question, the model completely fabricates a timeline, stating that Lucjan Brychczy was the coach before Zientara (1983–1985), which is inconsistent with any factual evidence supplied and temporally misaligned. This suggests the model introduces information outside the scope of the given context, undermining its reliability.
    \item \textbf{Misinterpretation of Predecessor-Successor Relationships:} The response to the fourth question incorrectly states that Zientara managed the team right before July 1969 (1968–1969), despite the clear evidence that Jaroslav Vejvoda was coach until June 1969 and Zientara only started in July 1969. This demonstrates poor reasoning when handling temporal predecessor-successor relationships, failing to factor transitional periods.
\end{itemize}

The findings necessitate a more rigorous and focused approach, even within flagship reasoning models such as DeepSeek-R1, to effectively address nuanced challenges in temporal reasoning, with specific emphasis on ensuring temporal referential consistency.

\begin{table*}[!t]
\centering
\small
\resizebox{.98\textwidth}{!}{
\begin{tabular}{l|l|p{10cm}}
\toprule

\multicolumn{3}{c}{\textbf{Wikkipedia Facts}}\\
\midrule
\multicolumn{3}{l}{Coaches of Legia Warsaw:}\\
\multicolumn{3}{l}{}\\
\multicolumn{3}{l}{Czech Republic Jaroslav Vejvoda (July 1966 – June 1969)}\\
\multicolumn{3}{l}{Poland Edmund Zientara (July 1969 – July 1971)}\\
\multicolumn{3}{l}{Poland Tadeusz Chruściński (July 1971 – May 1972)}\\
\multicolumn{3}{l}{Source: \url{https://en.wikipedia.org/wiki/Legia_Warsaw}}\\
\midrule
\multicolumn{3}{c}{\textbf{DeepSeek-R1 Responses}}\\
\midrule
\textbf{Direction}	&\textbf{Reference} &\textbf{Q/A} \\
\midrule
After & Chronological & \textbf{Question:} Who was the head coach of the team Legia Warsaw right after Edmund Zientara?\\
&&\textbf{Answer:} After Edmund Zientara left Legia Warsaw in 1980, the next head coach was Józef Zwierzyna, who took over the team in 1980 and managed Legia until 1981.
Here’s a quick timeline:
Edmund Zientara (1978–1980)
Józef Zwierzyna (1980–1981)
Would you like more details about Zwierzyna’s tenure or other Legia coaches?\\
\midrule
After & Absolute & \textbf{Question:} Who was the head coach of the team Legia Warsaw right after July 1971?\\
&&\textbf{Answer:} After July 1971, the head coach of Legia Warsaw was Edmund Zientara. He took over the team in July 1971 and managed Legia until the end of the 1972 season.
Zientara had previously played for Legia and later became a notable coach, leading the team to a Polish Cup victory in 1973 (after the 1972–73 season). His tenure was part of a transitional period for the club in the early 1970s.
Would you like more details about his coaching career or Legia's performance during that time? \\
\midrule
Before & Chronological & \textbf{Question:} Who was the head coach of the team Legia Warsaw right before Edmund Zientara?\\
&& \textbf{Answer:} The head coach of Legia Warsaw right before Edmund Zientara was Lucjan Brychczy.
Here’s the context:
Lucjan Brychczy served as Legia's head coach from 1983 to 1985.
Edmund Zientara took over in 1985 and coached the team until 1987.
Brychczy is a club legend, having also been a prolific striker for Legia in the 1950s–1970s before transitioning into coaching.
Would you like details on their tenures or achievements?

Who was the head coach of the team Legia Warsaw right before July 1969?\\

\midrule
Before & Absolute & \textbf{Question:} Who was the head coach of the team Legia Warsaw right before July 1969?\\
&&\textbf{Answer:} The head coach of Legia Warsaw right before July 1969 was Edmund Zientara.
He managed the team from 1968 to 1969 and was succeeded by Józef Walczak in July 1969.
Zientara was a notable figure in Polish football, both as a player and a coach, and he later had a successful managerial career, including leading the Poland national team in the 1980s.
Would you like more details about his tenure at Legia or his career in general? \\

\bottomrule
\end{tabular}
}
\caption{Qualitative analysis of DeepSeek-R1 on temporal placements across direction and references.}
\label{tab:deepseek}
\end{table*}

\section{Dataset and Quality}
\label{app:dataquality}

\paragraph{Dataset.} The formulation of queries using the event-event (L3) approach adheres to ten distinct templates, each pertaining to unique entities across six overarching categories of named entities, specifically: person, position, employer, political party, team, and school. The \newdata\ aligns with the original constituents of the source data (TEMPREASON), accompanied by comprehensive data statistics, as presented in Table \ref{tab:tempreasonl3}.

The TEMPREASON dataset, widely recognized as a standard benchmark in the field of temporal reasoning. Developed using multiple established predecessors, this dataset serves as the foundation for the novel TEMP-ReCon framework, consists of 13,014 unique queries across various temporal references, with a test set comprising 4,426 instances. The event-event category within the dataset spans six distinct entity types and encompasses ten knowledge base (KB) relations. Additionally, the dataset provides extensive time-period coverage, ranging from the year 634 to 2023. An event-event type of temporal query follows the structure in which, given a triplet of {subject, relation, and query-object}, the model is expected to identify answer-object, which is chronologically related to query-object in either forward or backward directions. Key phrases, such as `after' or `before,' indicate the anticipated direction of order.
\begin{table*}[!t]
\centering
\small
\resizebox{.98\textwidth}{!}{
\begin{tabular}{l|l|l|l}
\toprule
\textbf{KB Relation}&\textbf{\#Queries} &\textbf{Template} &\textbf{Anticipated General Entity Type}\\
\midrule
member of sports team & 2524 & Which team did <subject> play for before/after <object> ? & team\\
position held & 2538&Which position did <subject> hold before/after <object> ? & position\\
employer & 1991& Which employer did <subject> work for before/after <object> ?. & employer/company\\
political party &433 & Which political party did <subject> belong to before/after <object> ? & political party\\
head coach &1051 & Who was the head coach of <subject> before/after <object> ? & person\\
educated at &594 & Which school was <subject> attending before/after <object> ? & school\\
chairperson & 1881&Who was the chair of <subject> before/after <object> ? & person\\
head of government &1535 & Who was the head of the government of <subject> before/after <object> ? & person\\
head of state & 268& Who was the head of the state of <subject> before/after <object> ? & person\\ 
owned by &199 &Who was the owner of <subject> before/after <object> ? & person\\
\bottomrule
\end{tabular}
}
\caption{The structure of the event-event (L3) category from TEMPREASON is outlined in the following table. The first three columns are sourced from TEMPREASON.}
\label{tab:tempreasonl3}
\end{table*}

\paragraph{\newdata's Quality.} To ensure the proposed dataset's high quality, we first evaluated the translation success rate (TSR), which measures successful translations from source to target languages. An automated language detection method, langdetect\footnote{\url{https://pypi.org/project/langdetect}} applied to the entire test dataset revealed an remarkable TSR of 99.40\%. Additionally, we assessed translation quality using a back-translation\footnote{4 \url{https://translate.google.com/}} evaluation method \cite{Bajpai_Chakraborty_2025}. A sample of 100 queries from \newdata\ was randomly selected for each translated language, considering both absolute and chronological temporal references. These queries were back-translated into English and compared with their original texts. For quantitative assessment, we employed BERTScore-based F1-score and ChrF++ based metrics, resulting in a mean BERTScore (F1) of 71.49\%  and mean ChrF++ score of 45.49\% between the original and back-translated instances, focusing solely on queries across temporal references.

To evaluate inter-translator agreement, we utilized NLLB-200-1.3B alongside T5 on these 100 test samples. Translations were generated using NLLB for French and Romanian languages and then subsequently compared against T5 outputs to measure agreement at semantic and syntactical levels. This evaluation employed two metrics, BERTScore and BLEU-3, yielding mean translation agreement scores of 93.75 and 62.48, respectively across languages. The individual agreement scores for French were 93.38 (BERTScore) and 57.61 (BLEU-3), while Romanian achieved 94.13 (BERTScore) and 67.36 (BLEU-3). These results demonstrate high semantic alignment across both languages, with Romanian showcasing slightly stronger syntactic agreement compared to French. This highlights that T5 translation capability to achieve robust translation when compared to the recent benchmarks cush as NLLB.

\section{Hyperparameters}
\label{app:hyper}
In this study, we utilize the Python-based PyTorch\footnote{\url{https://pytorch.org}} library to implement all fine-tuning approaches. Additionally, to reproduce reinforcement learning-based baselines, we employed the Python-based library trlx\footnote{\url{https://github.com/CarperAI/trlx}}. The Hugging Face platform was utilized to stack open-source models, including the base model LLaMA3.1-8B, which served as the foundation for experimentation. For the execution of these experiments, an NVIDIA A100 GPU\footnote{\url{https://www.nvidia.com/en-in/data-center/a100/}} with 80GB of memory was employed. We set the values of lora\_r and lora\_alpha at 16 and 32, respectively, for the LoRA adapter and fine-tuned the \newmodel\ for 20 epochs across all supervised fine-tuning methods. We convert the \newdata's training data into the instruction format proposed by \cite{alpaca} (an instance is presented in Figure \ref{fig:itprompt}). Furthermore, we applied 8-bit quantization to optimize computational efficiency and accelerate training. An alpha value of 0.66, as recommended by the authors of CoTSeLF, was utilized to run the baseline. Benchmarking temporal referential consistency across models is performed using the default parameter settings of the base models for temperature, top\_p, and top\_k, along with num\_beams set to 2 and max\_new\_tokens set to 30. During fine-tuning, we set the max token cut\_off limit to 256.

We selected 2000 random parallel instances (\~15\%) from the training data to construct the paired sentences dataset for the consistency task for CoTSeLF baseline. A positive pair across temporal references and an antagonist pair each occur within the temporal framework. 

\section{Prompts}
\label{App:prompts}
\begin{figure*}[!t]
\centering
\subfloat[English]{\includegraphics[width=01\textwidth]{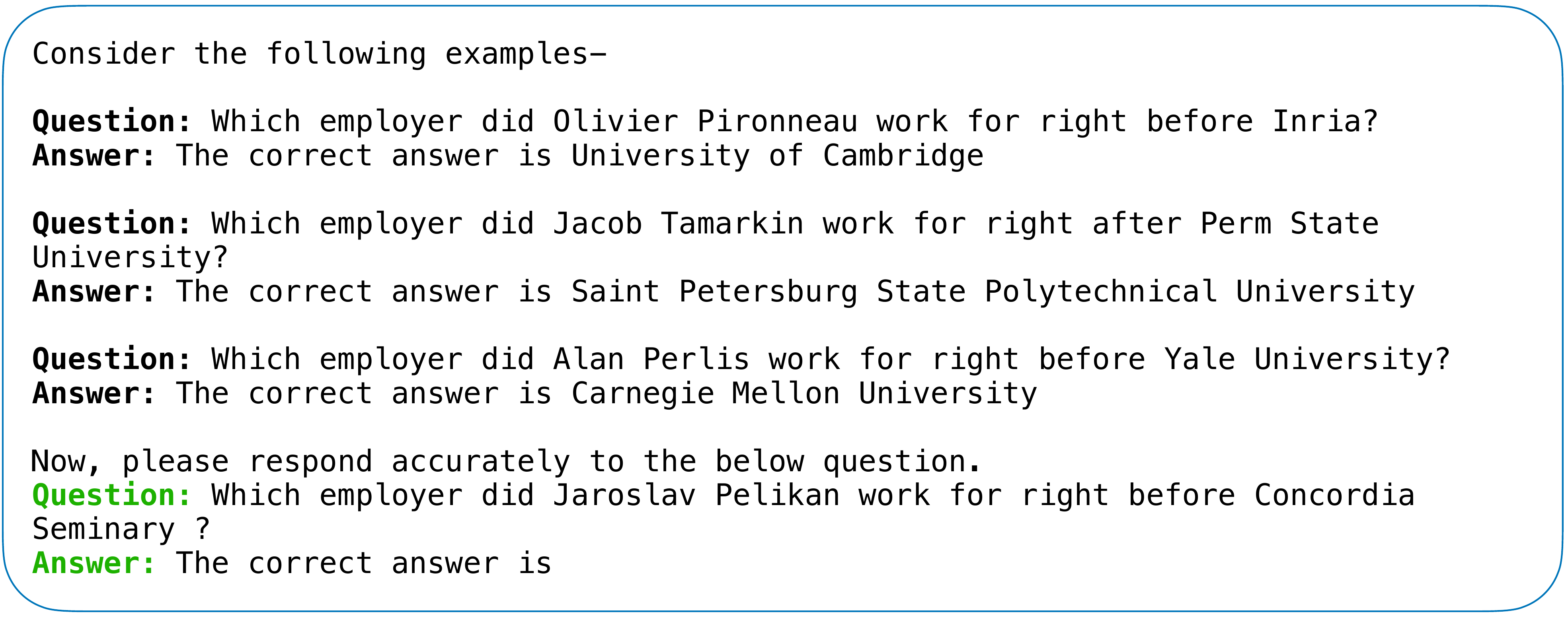}}\\ 
\subfloat[French]{\includegraphics[width=1\textwidth]{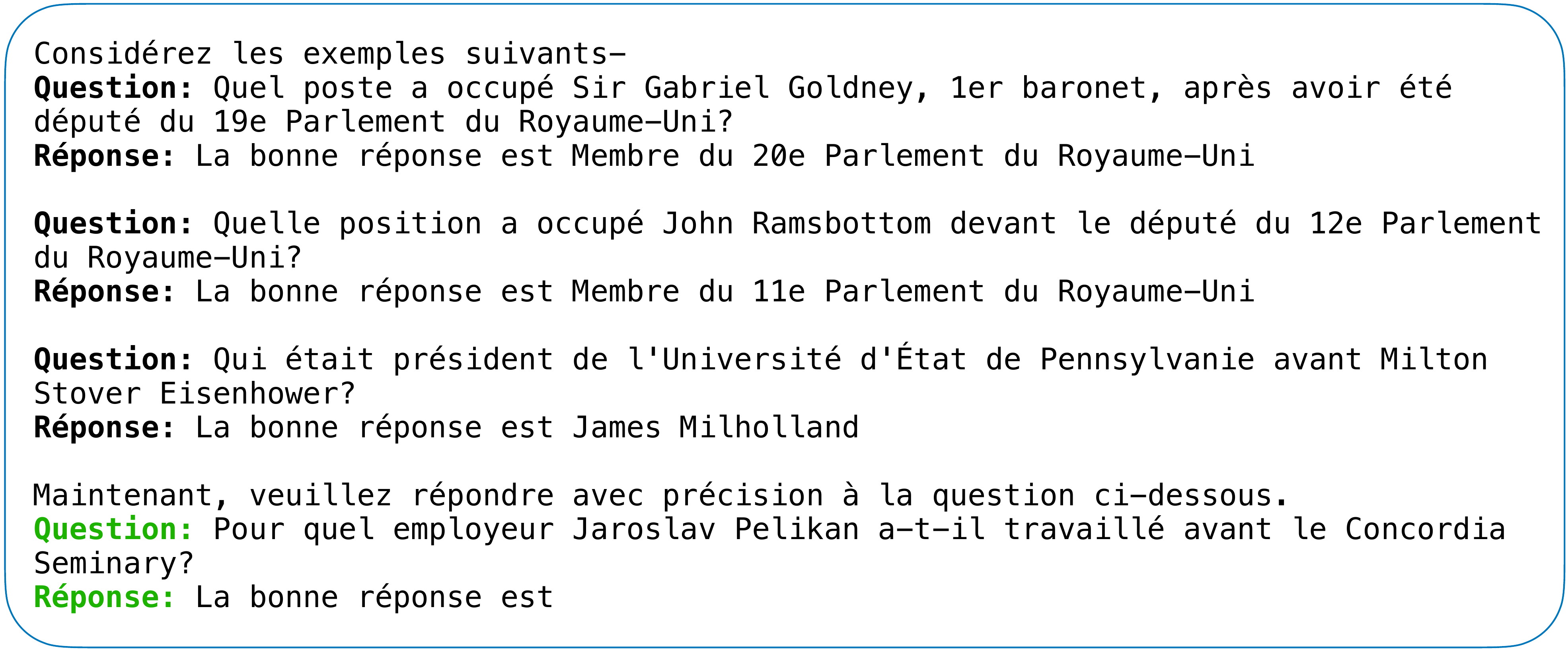}} \\
\subfloat[Romanian]{\includegraphics[width=1\textwidth]{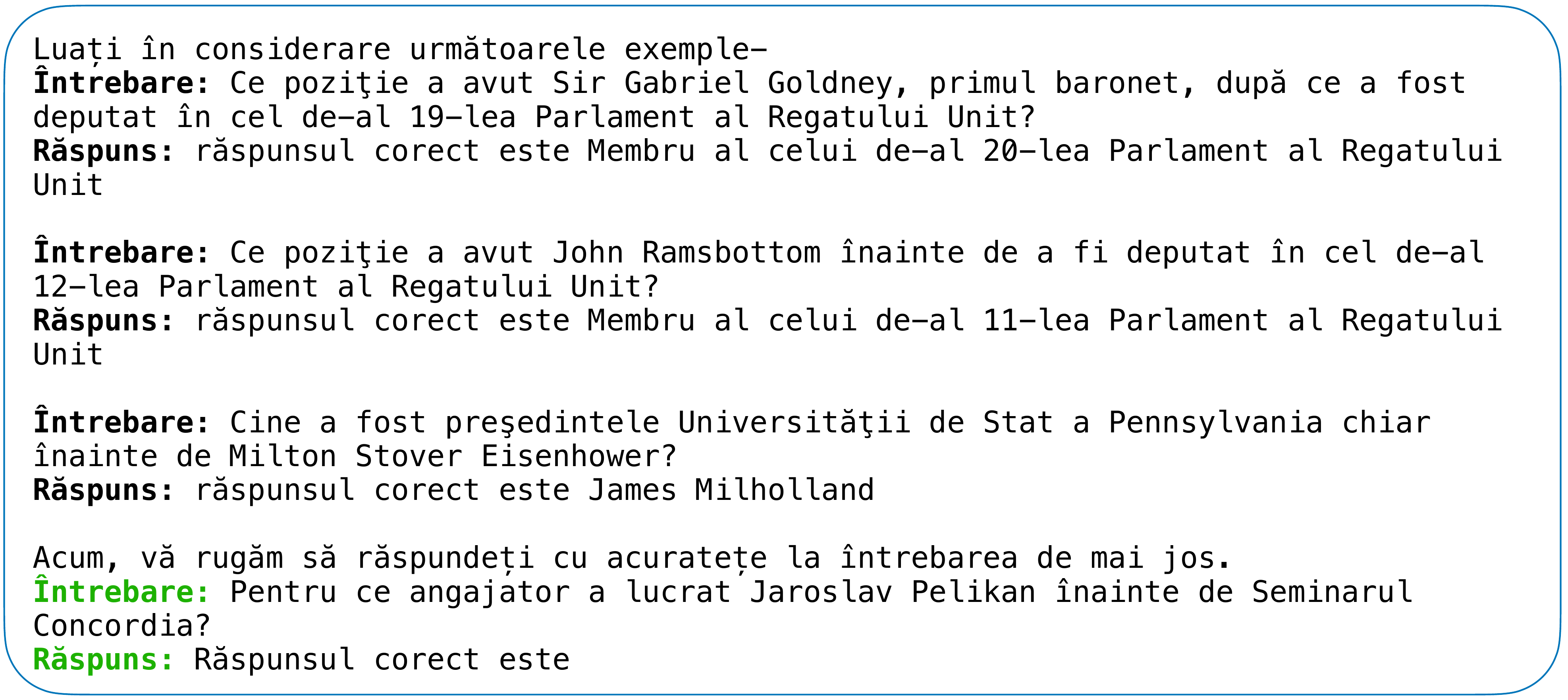}} 
\caption{A three-shot prompt that is used to assess temporal referential consistency of LLMs in experiments involving in-context learning (ICL) methods.}
  \label{fig:fewshot}
\end{figure*}

\begin{figure*}[!t]
\centering
\includegraphics[width=0.95\textwidth]{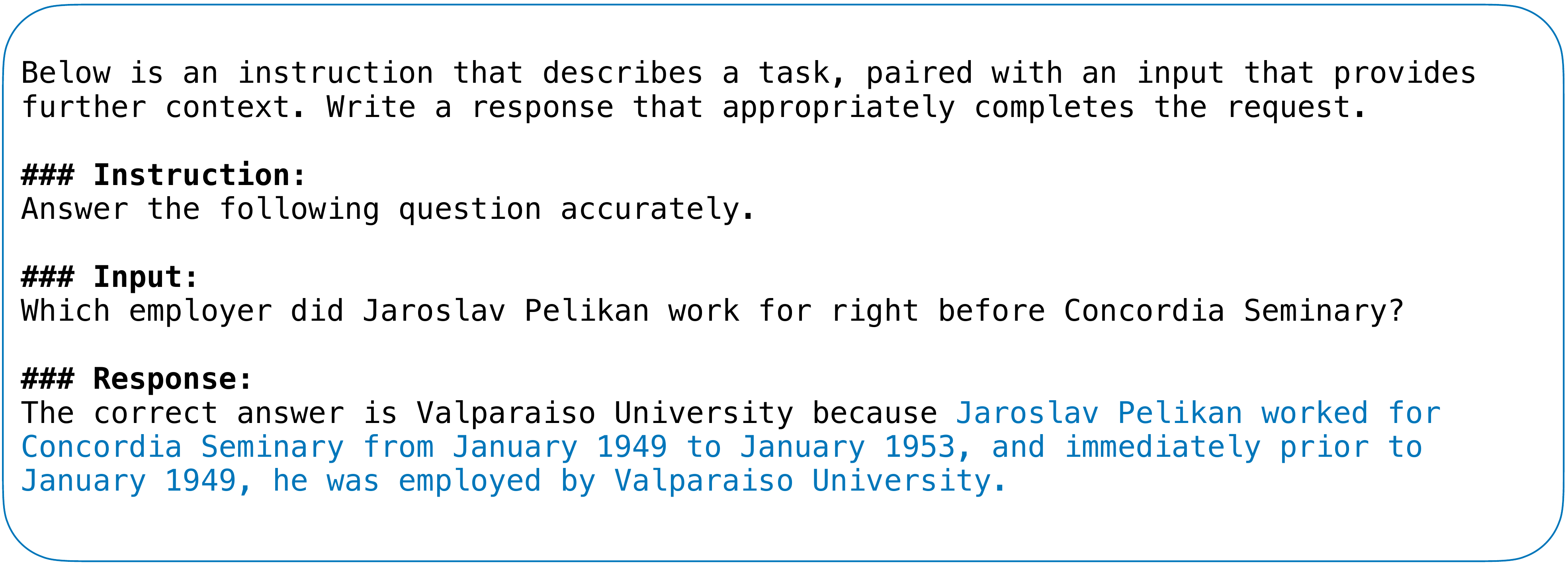}
  \caption{An example from the training data for chronological temporal reference-based queries, along with the time-oriented reasoning path used in \newmodel's instruction tuning.}
  \label{fig:itprompt}
\end{figure*}

An example of a few-shot prompt utilized in in-context settings is illustrated in Figure \ref{fig:fewshot}, encompassing various languages. Furthermore, a representative instance of the training data employed during the fine-tuning process is depicted in Figure \ref{fig:itprompt}.

\section{Commentaries}
\label{App:commentary}

\paragraph{Domain Adaptation.} Temporal referential consistency can be adapted across various domains, particularly in time-sensitive fields such as finance and law. For instance, queries such as \textit{"How did the US economy perform during 2008-2016?"} and \textit{"How did the US economy perform under Barack Obama?"} are semantically identical yet employ differing temporal references: a specific time period (2008-2016) versus a notable individual (Barack Obama). It is imperative that responses generated by models exhibit consistency in addressing such queries across diverse temporal references. 

In the context of finance, the accurate and consistent retrieval of economic indicators and financial data is fundamental for informed decision-making. Inconsistencies in responses provided by language models (LLMs) to diverse temporal reference inquiries regarding the same subject can result in suboptimal investment strategies, misinformed clientele, or inaccuracies in regulatory filings. Similarly, in the legal domain, variations in responses to queries that reference the same legal principle or case but through different temporal references may evoke confusion and increase the risk of legal malpractice. Consequently, establishing temporal referential consistency in time-sensitive domains is critical for enhancing accuracy, reliability, and risk mitigation. It is anticipated that future research will extend or adapt the findings of the current study across various domains.
\vspace{-2mm}
\paragraph{Applicability For Closed-Source Models.} The commercial models, such as GPT-4 and Claude-3, exhibit suboptimal performance in terms of temporal referential consistency. This limitation necessitates the development of strategies, including the proposed \newmodel\ or analogous enhancements, aimed at improving this aspect of temporal consistency. The \newmodel\ introduces an alignment-based methodology that seeks to unify reasoning paths across various temporal references through the refinement of model parameters. Due to the restrictions placed on direct architectural access in models like GPT-4 and Claude-3, it is imperative that such methodologies demonstrate their effectiveness solely on open-source models. 

Considering that the predominant architecture for contemporary commercial large language models is based on decoder-based transformers, the \newmodel\ approach can be effectively applied during either the fine-tuning stage or through reinforcement learning from human feedback, serving as an extended objective alongside existing methodologies. Furthermore, closed-source models can benefit from a similar methodological framework implemented via non-gradient update approaches that do not necessitate access to the model's parameters. However, we have chosen to adopt a fine-tuning-based methodology, as context-based approaches have been shown to yield suboptimal performance compared to fine-tuning methodologies.

\end{document}